\documentclass{article} % For LaTeX2e
\usepackage{conference}

\usepackage[utf8]{inputenc}
\usepackage{graphicx}
\usepackage{makecell}
\usepackage{inconsolata}
\usepackage{amsmath,bm,amssymb}
\usepackage{microtype}
\usepackage{hyperref}
\usepackage{url}
\usepackage{booktabs}
\usepackage{todonotes}
\usepackage{float}
\usepackage{comment}
\usepackage{nameref}
\usepackage{wrapfig}
\usepackage[most]{tcolorbox}
\usepackage{lipsum}
\usepackage{xcolor} 
\usepackage{caption}
\usepackage{pifont}
\usepackage{tikz}
\usepackage{multirow} 
\usepackage{wrapfig}
\usepackage{subcaption}

\definecolor{softGray}{HTML}{FFFFFF}
\definecolor{deepBlue}{HTML}{003366}
\definecolor{darkGray}{HTML}{333333}
\definecolor{highlightBackground}{HTML}{E0DBE8}
\newcommand{\highlight}[1]{\colorbox{highlightBackground}{#1}}

\newtcolorbox{stage1}{
    colback=softGray, % Soft Gray
    colframe=deepBlue, % Deep Blue
    coltext=darkGray, % Dark Gray
    title=Instruction for Generating Explanation,
    fontupper=\fontsize{8pt}{1pt}\selectfont,
}

\definecolor{nb}{HTML}{006EB8}
\usepackage{hyperref}
\hypersetup{
  colorlinks   = true,    % Colours links instead of ugly boxes
  urlcolor     = nb,    % Colour for external hyperlinks
  linkcolor    = nb,    % Colour of internal links
  citecolor    = nb      % Colour of citations
}

\usepackage[ruled,lined,linesnumbered]{algorithm2e}
\newcommand{\methodname}{{\tt{LMExplainer}}}
\definecolor{lightgreen}{HTML}{D5DE56}
\definecolor{lightblue}{HTML}{D1E9EC}
\definecolor{black}{HTML}{000000}

\newcommand{\zc}[1]{{\color{black} #1}}

\title{LMExplainer: Grounding Knowledge and Explaining Language Models}

\colmfinalcopy
\author{Zichen Chen \\
UC Santa Barbara\\
\texttt{\{zichen\_chen\}@ucsb.edu} \\
\And
Jianda Chen, Yuanyuan Chen, Han Yu \\
Nanyang Technological University \\
\texttt{\{jianda001,yuanyuan.chen,han.yu\}@ntu.edu.sg} \\
\And
Ambuj Singh, Misha Sra \\
UC Santa Barbara\\
\texttt{\{ambuj,sra\}@ucsb.edu} \\
}

\begin{document}

\maketitle

\begin{abstract}

Language models (LMs) like GPT-4 are important in AI applications, but their opaque decision-making process reduces user trust, especially in safety-critical areas.
We introduce \methodname{}, a novel knowledge-grounded explainer that clarifies the reasoning process of LMs through intuitive, human-understandable explanations. By leveraging a graph attention network (GAT) with a large-scale knowledge graph (KG), \methodname{} not only precisely narrows the reasoning space to focus on the most relevant knowledge but also grounds its reasoning in structured, verifiable knowledge to reduce hallucinations and enhance interpretability. 
% precisely narrows the reasoning space, focusing on the most relevant knowledge and explains how LMs arrive at their conclusions. 
\methodname{} effectively generates human-understandable explanations to enhance transparency and streamline the decision-making process. Additionally, by incorporating debugging into the  explanation, it offers expertise suggestions that improve LMs from a developmental perspective. 
Thus, \methodname{} stands as an enhancement in making LMs more accessible and understandable to users.
We evaluate \methodname{} on benchmark datasets such as CommonsenseQA and OpenBookQA, demonstrating that it outperforms most existing methods. By comparing the explanations generated by \methodname{} with those of other models, we show that our approach offers more comprehensive and clearer explanations of the reasoning process.
\methodname{} provides a deeper understanding of the inner workings of LMs, advancing towards more reliable, transparent, and equitable AI. 

\end{abstract}

\section{Introduction}

Language models (LMs) have recently attracted significant attention due to their impressive state-of-the-art (SOTA) performance on various natural language processing (NLP) tasks \citep{Zhao2024,liu2023pre,zhao2023survey,weifinetuned,zhou2022learning}. These tasks include language translation \citep{barrault2023seamless}, text generation \citep{dekoninck2024controlled}, and text classification \citep{liu-etal-2023-enhancing}, among others. One of the main advantages of LMs is their ability to capture the nuances and the complexity of human languages.

% \begin{figure}[t]
%   \begin{center}
%     \includegraphics[width=0.48\textwidth]{figure/intro.pdf}
%   \end{center}
%   \caption{\methodname{} demystifies the decision-making process of LMs for better human understanding. It includes a graph surrogate for structural reasoning, attention-based interpretation for decision rationales, and an explanation generator that provides explanations of \textit{``why-choose''} and \textit{``why-not-choose''} to bridge the gap between LMs and human understandability.}
%   \label{fig:intro}
%   \vspace{-10pt}
% \end{figure}

However, a major limitation of LMs is a lack of interpretability \citep{meng2022interpretability}. It is often difficult to provide explanations about their ``black box'' decision-making processes. LMs use techniques such as attention mechanisms, which allow them to focus on specific parts of the input data when making decisions \citep{ devlin2019bert,liu2019roberta,mrini-etal-2020-rethinking}. These mechanisms can be difficult for people to understand, as they produce abstract and non-transparent internal learning representations \citep{jain2019attention}. %But these mechanisms can elude human intuition, and produce internal learning representations that are abstract and non-transparent to humans \citep{jain2019attention}.
For example, a model embedding might capture relationships and meanings as a result of passages through millions of neurons. However, such meanings might not be immediately apparent to humans. This lack of interpretability poses a challenge to mission critical domains (e.g., healthcare \citep{jung2023essential} and online education \citep{farrow2023possibilities}) as it hampers users' trust on the responses made by the models.
%Improving model interpretability has the added benefit of addressing issues such as fairness, privacy, and safety. Methods that explain the behaviors of LMs can help overcome the black-box nature of neural networks.

Due to the opaque nature of LMs, a promising approach for explaining how they work is by generating explanations on a more transparent surrogate (e.g., a knowledge graph (KG)).
\citep{geng2022path} leverages a KG as a submodel to ground the explainability of LM-based recommendations.
Such methods provide insights into how to interpret the complex model by translating it into more comprehensible counterparts. Attention-based explanations have also gained significant attention. For example, \citep{vig-2019-multiscale} proposes a visualization method for attention in the LM, enhancing our understanding of how these models allocate focus across input tokens. However, \citep{zini2022explainability} pointed out that attention is not equal to explanation. Individual token representations are not enough. A surrogate that maps tokens to specific knowledge elements that align with the reasoning process of the LM is imperative.

% In this paper, we explore the potential of using explanations to serve two purposes (Figure \ref{fig:intro}): \textbf{1) helping humans in understanding the model}, and \textbf{2) enhancing the model's understanding of the task at hand through interpretation during the explanation process}. 

To address the limitations of current approaches, we propose the~\methodname{} approach, a novel method for \zc{explaining the mysteries behind LM's decisions. Unlike other methods that focus on single token or attention, \methodname{} treats reasoning as an integrated process. }
% explaining the responses made by LMs.
It is designed to efficiently locate the most relevant knowledge within a large-scale KG via the graph attention neural network (GAT) \citep{velickovic2018gat} to extract key decision signals reflecting the rationale behind the responses made by LMs. \zc{Through the graph framework, we narrow the reasoning space and reduce the potential noise, thereby improving the inference accuracy and providing a transparent structure for explaining model's reasoning behavior. Our method achieves grounded explanations by anchoring the reasoning process in verifiable knowledge, which reduces hallucinations. Furthermore, we incorporate a debugging process to quantitatively understand the generated explanations and provide suggestions for future enhancements.}

We experimentally evaluate~\methodname{} on the question-answering (QA) task using the CommonsenseQA~\citep{talmor2019commonsenseqa} and OpenBookQA~\citep{mihaylov2018can} datasets. The results demonstrate that~\methodname{} outperforms most LM+KG QA methods and large LMs (LLMs) on CommonsenseQA and OpenBookQA.
\zc{Furthermore, we show that~\methodname{} surpasses prior explanation methods by providing insights of the reasoning behind a LM's decision in a human-understandable form and LM development suggestions. We open up new possibilities for AI systems to be not only more transparent and reliable but also equitable.}
To the best of our knowledge,~\methodname{} is the ﬁrst work capable of leveraging graph-based knowledge in generating natural language explanations on the rationale behind LM behaviors.
\section{Related Work}

Post-hoc explanation methods have attracted significant attention in NLP research in recent years.
\citet{ribeiro2016should} proposed LIME, which generates explanations by approximating the original model with a local sample and highlights the most important features.~\citet{guidotti2018local} extended it with a decision tree classifier to approximate deep models. However, they cannot guarantee that the approximations are accurate representations of the original model due to inherent limitations of decision trees.~\citet{thorne2019generating} generate concepts of classifiers operating on pairs of sentences, while~\citet{yu2022towards} generate \emph{aspects} as explanations for search results.~\citet{kumar-talukdar-2020-nile} used positive labels to generate candidate explanations, while~\citet{chen2021kace} used contrastive examples in the form of ``why A not B'' to distinguish between confusing candidates.  Different from prior work, we integrate reasoning features and concepts into \methodname{} to explain LM behaviors.

Recently, language models (LMs) such as RoBERTa~\citep{liu2019roberta}, Llama~\citep{touvron2023llama} and GPT-4~\citep{OpenAI2023GPT4TR}  have achieved impressive results.
However, these models lack interpretability, which can hinder their adoption in mission critical real-world applications.
Previous interpretable frameworks \citep{ribeiro2016should,mukund2017integratedgrad,Smilkov2017SmoothGradRN,ding2021saliencyNLP,swamy2021interpreting} can be applied to LMs. However, they often rely on approximations and simplifications of the original models, which can result in discrepancies between the model behaviours and the explanations.
In contrast, \methodname{} explains LMs by illustrating the model reasoning process.

KGs are increasingly adopted as a means to improve the interpretability and explainability of LMs \citep{huang-etal-2022-deer, yasunaga2021qagnn, huang2019explainable, liu2019knowledge}. KGs are structured representations of knowledge, and can be used to capture complex semantic relationships that are difficult to represent in traditional LMs \citep{ji2021survey}.
\cite{ZHAN2022107612} retrieves explainable reasoning paths from a KG and uses path features to predict the answers.
\cite{yasunaga2021qagnn} integrates the KG into the model, enabling the model to reason over structured knowledge and generate more interpretable predictions.
However, these explanations can be inconsistent and accurate representations of the model reasoning process. In addition, they are difficult for humans to understand as they are being represented in a graph-based format.
By drawing upon insights from prior works, \methodname{} employs graph embedding to generate explanations to address these limitations.
\section{The Proposed \methodname{} Approach}
\begin{figure*}[h]
  \begin{center}
    %\framebox[4.0in]{$\;$}
    %\fbox{\rule[-.5cm]{0cm}{4cm} \rule[-.5cm]{4cm}{0cm}}
    \includegraphics[width=1\textwidth]{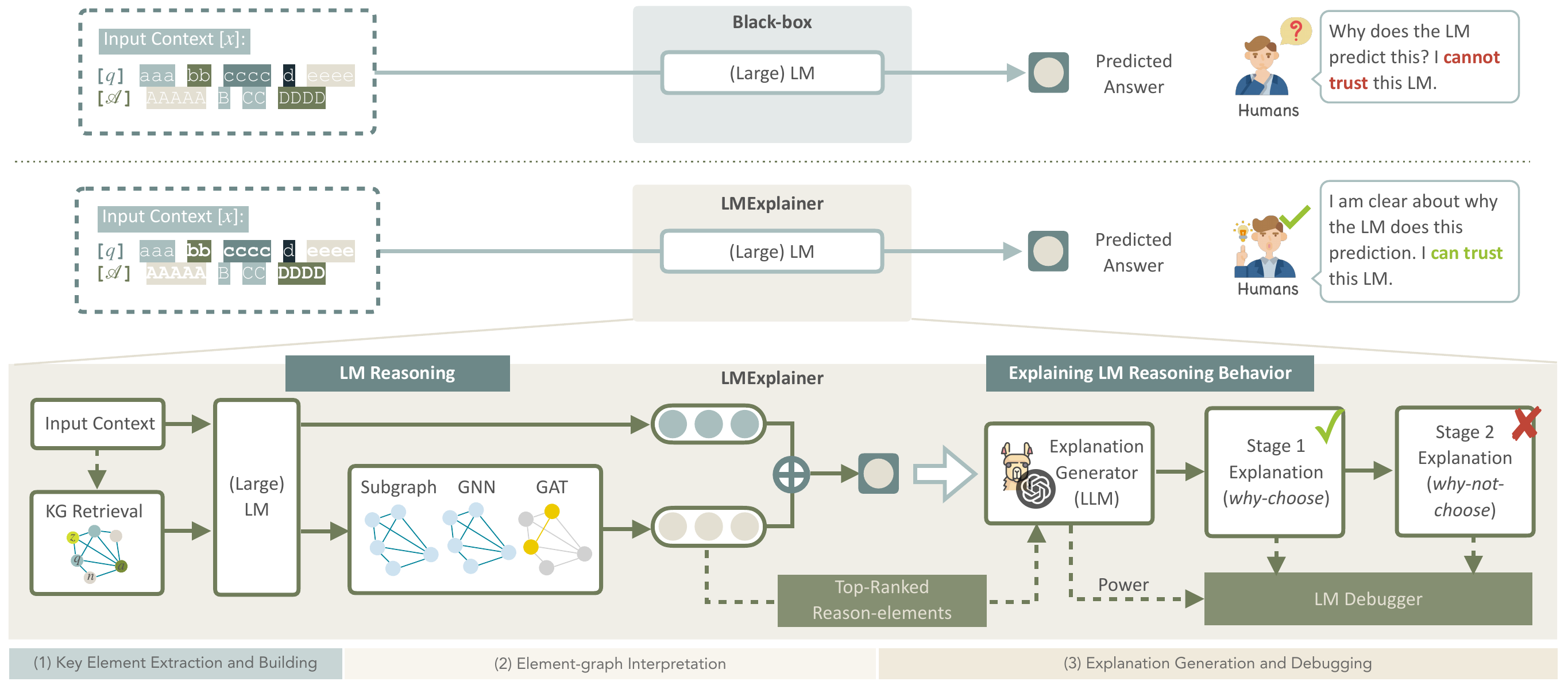}
  \end{center}
  \caption{\zc{The~\methodname{} architecture. \methodname{} addresses the lack of transparency in black-box LMs by providing interpretable explanations for their reasoning behavior. Given an input context, \methodname{} retrieves relevant knowledge from a KG, integrates it with LM embeddings to construct a surrogate subgraph, and interprets it using graph attention to identify key reason-elements contributing to the LM's decision. Based on these elements, LMExplainer generates explanations clarifying why the LM chose the predicted answer and ruled out alternatives. The LM Debugger further enhances user trust by evaluating the faithfulness, completeness, minimality, and accuracy of the generated explanations.} }
  % The~\methodname{} architecture. Given an input content $z$, we first generate language embeddings using a pre-trained LM. Simultaneously, it retrieves relevant knowledge from a KG to construct a subgraph. The language embeddings and subgraph are then combined to obtain GNN embeddings. This combined representation is then passed through a GAT to obtain the attention. The attention serves two purposes. Firstly, it weighs the importance of the GNN embeddings and is used with the language embeddings for the final prediction. Secondly, they are used to generate explanations by highlighting the most important parts of the reasoning process. }
  \label{fig:archi}
  \vspace{-10pt}
\end{figure*}

The~\methodname{} architecture is shown in Figure~\ref{fig:archi}. It consists of three main steps: \textbf{(1) key element extraction and building} (Section~\ref{sec:extraction}), \textbf{(2) element-graph interpretation} (Section~\ref{sec:interpretation}), and \textbf{(3) explanation generation and debugging} (Section~\ref{sec:explanation}). 
% In the first step, we extract the relevant elements from the input data and the knowledge retrieved from the KG, and build an element-graph representation. In the second step, we leverage GAT to interpret the element-graph and identify the \emph{reason-elements} behind LM predictions. In the third step, we design an instruction-based method to generate human-understandable explanations of the decision-making process based on the identified \emph{reason-elements}.
\methodname{} is flexible and applicable to a range of LMs (e.g., RoBERTa \citep{liu2019roberta}, GPT-2 \citep{radford2019language}, and Llama-2 \citep{touvron2023llama-2}).

\subsection{Task Definition}
We define the task of generating reasoning-level explanations for inferences made by LMs. 
% As an example, we use a QA task. 
\zc{Given a LM $f_{\text{LM}}$, input $x$, and a predicted output $y'$, the goal is to generate a natural language explanation $E$ that elucidates the reasoning process behind the prediction. The task can be expressed as:
    \begin{equation}
        E \gets \text{GenerateExplanation}(f_{\text{LM}}, x, y')
    \end{equation}
As an example, we use a QA task, where input $x$ consists of question $q$ and a set of answer choices $\mathcal{A} = {a_1, a_2, \ldots, a_n}$, the LM $f_{\text{LM}}$ predicts an answer $y' \in A$. The explanation $E$ consists of $E_{0}$ and $E_{1}$. $E_{0}$ is for why $f_{LM}$ chooses $y'$ and $E_{1}$ is for why $f_{LM}$ does not choose other choices $\mathcal{A} \setminus \{ y'\}$.}

% Given an LM $f_{LM}$ with input question $q$, answer choice set $\mathcal{A}$ and predicted answer $y' \in \mathcal{A}$, the goal is to generate an explanation $E_{0}$ for why $f_{LM}$ chooses $y'$ and an explanation $E_{1}$ for why $f_{LM}$ does not choose other options $\mathcal{A} \setminus \{ y'\}$. This task can be expressed as:
% \begin{equation}
%   (E_{0},E_{1}) \gets \textit{GenerateExplanation}(f_{LM}, q,\mathcal{A}, y').
% \end{equation}

\subsection{Key Elements Extraction and Building} \label{sec:extraction}
\zc{The reasoning patterns of LMs can be traced by key elements that reveal their decision-making logic \citep{huang2023towards}.
To capture these complex reasoning signals, we propose a more structured and transparent surrogate method that effectively maps the LM's reasoning process into an interprerable representation. 
We first tokenize the input $x$ and let $z$ denote the resulting tokens.}
% To capture these essential elements, we first tokenize a set of sentences $\{q\} \cup \mathcal{A}$ into tokens $\{x_1, x_2,\ldots, x_n\}$. Let $z$ denote this set of resulting tokens. 
The tokens $z$ are then used to construct a multi-relational graph, following the approach from~\citeauthor{yasunaga2021qagnn}.
\zc{We retrieve the $L$-hop neighborhood graph $G_k$ of $z$ from a large-scale knowledge graph $\mathcal{G}$, such as ConceptNet \citep{speer2017conceptnet}. To narrow down the reasoning space and align it with the LM's own reasoning patterns, we construct a relevant subgraph of $\mathcal{G}_k$, referred to as the \textit{element-graph} $\mathcal{G}_e$, by leveraging the LM's knowledge to guide the pruning process. This pruning step is crucial for two reasons. First, it effectively focuses on the most informative and relevant elements, ensuring that the selected elements are consistent with the LM's decision-making process. Second, it captures and reflects the LM's capabilities, preferences, and potential biases, making them transparent and interpretable.

For each node $v$ in $\mathcal{G}_k$, we define a relevance score $s_v$ that measures the importance of the node with respect to the input text $x$. The relevance score is computed using the LM's probability function $f_{\text{prob}}$ as follows:
\begin{equation}
    s_v = f_{\text{prob}}(z_{\text{emb}}, v_{\text{emb}})
\end{equation}
where $z_{\text{emb}}$ and $v_{\text{emb}}$ are the embeddings of the input tokens $z$ and the node $v$, respectively, obtained from the LM. The probability function $f_{\text{prob}}$ takes the concatenated embeddings as input and outputs a scalar value indicating the relevance of the node to the input text. We select the top-$K$ nodes based on their relevance scores to construct the \textit{element-graph} $\mathcal{G}_e$. The procedure for constructing the \textit{element-graph} is outlined in the Appendix (Algorithm \ref{algo:subgraph}).

}

% Firstly, the $L$-hop neighbor $G_k$ of $z$ is extracted from ConceptNet \citep{speer2017conceptnet} to integrate external knowledge, following the approach from ~\cite{feng2020scalable}. However, $G_k$ can still contain a large number of edges, which lead to a huge reasoning space. Our main goal is therefore to construct a relevant sub-graph of $G_k$, referred to as the \textit{element-graph} $G_e$.
% This allows us to identify essential elements that play a key role, and analyze the relations among them.
% We integrate the embedding from LMs to guide the pruning for $G_k$. Specifically,
% for every node $v$ in $G_k$, we define an associated score for pruning purposes, which is expressed as:
% \begin{equation}
%   v_{score} = f_{prob}(z_{emb},  v_{emb}),
% \end{equation}
% where $f_{prob}$ is the probability computation function of the pre-trained LM, $z_{emb}$ and $v_{emb}$ are the embeddings derived from textual representations of $z$ and $v$ respectively, are concatenated to $f_{prob}$. The score captures the correlation between the node $v$ and input content $z$, and is used to remove irrelevant nodes. We select the top $K$ nodes based on their scores. The resulting pruned graph is denoted by $G_e$, which is referred to as the \textit{element-graph}. We outline the procedure for constructing the \textit{element-graph} in the Appendix (Algorithm \ref{algo:subgraph}).

 \subsection{Element-Graph Interpretation}  \label{sec:interpretation}
Given an element-graph $G_e$, we follow~\cite{yasunaga2021qagnn} to extract the representation for graph reasoning. The method leverages the GAT~\citep{velickovic2018gat} to preserve the structure and context of the input through the connections between the nodes.
\citeauthor{velickovic2018gat} use the graph attention operation to take a set of node features as input and output a corresponding set of new node features. Formally, the input to the $k \text {th}$ attention layer is denoted as $\boldsymbol{h}_k=\{h_{k1},h_{k2}, \ldots,h_{kN}\}$, where $h_{kj} \in \mathbb{R}^F$ is the intermediate feature for node $v_j$, $F$ is the input feature size and $N$ is the number of nodes in the graph. The attention layer outputs a new set of corresponding node features, $\boldsymbol{h}_{k+1}=\{h_{k+1,1}, h_{k+1,2},\ldots, h_{k+1,N}\}$ with $h_{kj} \in \mathbb{R}^{F^{'}}$.

A parameterized transformation $m: \mathbb{R}^F \xrightarrow{} \mathbb{R}^M $ is first applied to $\boldsymbol{h}_k$ to generate the transformation $m(\boldsymbol{h}_k)$. A parameterized self-attention mechanism $a: \mathbb{R}^F \times \mathbb{R}^F \xrightarrow{} \mathbb{R}$ is then used to obtain attention scores on $\boldsymbol{h}_k$. To retain structural information within the graph, attention scope for node $v_i$ is limited to nodes in its 1-hop neighborhood which is denoted as $\mathcal{N}_i$. Furthermore, the attention scores are normalized over the neighborhood $\mathcal{N}_i$ to generate attention coefficients:
\begin{equation}
  \alpha_{ij} = \frac{\exp(a(h_{ki},h_{kj}))}{\sum \limits_{v_l \in \mathcal{N}_i } \exp(a(h_{ki},h_{kl}))}.
  \vspace{-2mm}
\end{equation}

The output feature $h_{k+1,i}$ is an attentive linear combination of neighboring features with an optional activation:
\begin{equation}
  h_{k+1,i} = \sigma( \sum \limits_{v_j \in \mathcal{N}_i }  \alpha_{ij} m( h_{kj}))
    \vspace{-2mm}
\end{equation}

We build the graph reasoning network based on the above graph attention operation. Specifically, we employ a parameterized MLP $f_m$ for feature transformation. This MLP $f_m$ explicitly associates the node $v_i$ with its neighboring nodes $v_j \in \mathcal{N}_i$ by processing the feature $h_{ki}$, the recorded node type $u_i$ of node $v_i$ and the recorded relation types $r_{ij}$ to $v_j$, all of which are sourced from the element-graph. 
The attention scores $\alpha_{ij}$ are computed using another parameterized MLP that takes features $h_{ki},h_{kj}$, node and relation types $u_i,r_{ij}$ and node scores of $v_i$ and $v_j$ as input. The detailed information can be found in the Appendix~\ref{sec:graph}.

The output activation is implemented as a third 2-layer parameterized MLP $f_\sigma$ and the output features are thus obtained by:
\begin{equation}\label{eq:h_ve}
  h_{k+1,i} = f_\sigma ( \sum \limits_{v_j \in \mathcal{N}_i }  \alpha_{ij} m(h_{kj},u_i,r_{ij}))+ h_{kj},
    \vspace{-2mm}
\end{equation}
where the output feature size is the same as the input feature size.
The initial input features $\boldsymbol{h}_0$ is obtained by a linear transformation of node embeddings $v_{emb}$.

\subsubsection{Learning and Inference} \label{sec:inference}
In our task, each question $q$ is associated with a set of answer choices $\mathcal{A}$, with only one being the correct answer. We leverage the information from the LM embedding and the node embedding from the element-graph. Specifically, we define the probability of choosing an answer as $P(a|q)\propto \exp(\textit{MLP}(\mathbb{H}^{LM}, \boldsymbol{h}_K , \boldsymbol{\alpha}_K))$, where $\boldsymbol{h}_K$ is the output features and $\boldsymbol{\alpha}_K$ is the last-layer attention coefficients of a K-layer graph reasoning network given $G_e$ as input, and $\mathbb{H}^{LM}$ is the representation embedding from LM. The corresponding nodes (i.e., the \emph{reason-elements}) in $G_e$ are used to generate textual explanations about the decision-making process of the LM.
We optimize the model by minimizing the cross-entropy loss.

\subsection{Attention-aware Explanation Generation and Debugging} \label{sec:explanation}
% In prior work, ~\cite{chen2021kace} proposed a counterfactual explanation generator that pairs input text with counterfactual examples to fine-tune LMs to generate explanations in the form of ``why A and not B''. However, this approach does not provide a complete explanation of the LM decision-making process.

% In contrast, 
The \methodname{} explanation generator consists of three steps: (1) explanation component extraction, (2) instruction-based explanation generation and \zc{(3) enhancing explanation comprehension with LLM-based debugging. }

\subsubsection{Explanation Component Extraction}
We first extract the key components that are essential to the LM decision-making process. These key components consist of the final answer, \emph{reason-elements} and the attention $\alpha$. The final answer and \emph{reason-elements} are used to trace the important explanation nodes. The attention is used to sort the nodes and select the top $w$ nodes most relevant to the decision. 
% Each node represents an element, so we have $w$ most important components for the explanation. 
We use $\mathcal{Q}$ to represent the set of extracted key components. 
% The output, $E$, is a natural language explanation. 
We outline the procedure to interpret the \emph{element-graph} and extract the \emph{reason-elements} in the Appendix (Algorithm \ref{algo:element}).

\subsubsection{Instruction-based Explanation Generation}
We integrate the key component set $\mathcal{Q}$ into our instruction-based explanation generator. 
\zc{We leverage a set of predefined structures to guide the explanation generation. }
% To guide the generation of explanations, we leverage a set of predefined structures, including the input $q$ and $\mathcal{A}$, model predicted output $y'$, the trigger sentences, and the extracted key components $\mathcal{Q}$.
The \methodname{} explanation generation involves two stages: (1) \emph{why-choose} for explaining why the model chose the specific answer, and (2) \emph{why-not-choose} for explaining why the model did not choose the other explanations. 
\zc{In the \emph{why-choose} stage, we use instructions in the form of ``\textbf{Basis:} [TASK\_TYPE], \textbf{Input:} [$q$, $\mathcal{A}$], Output: [$y'$, $\mathcal{Q}$], \textbf{Explanation (Stage 1):} [$y'$]''. The \emph{why-choose} explanation is denoted as $E_0$. In the \emph{why-not-choose} stage, we use instructions in the form of ``\textbf{Explanation (Stage 2):} [$E_0$, $\mathcal{A}\setminus\{y'\}$]''. }
\zc{\textit{Basis}, \textit{Input}, \textit{Output}, \textit{Explanation (Stage 1)}, and \textit{Explanation (Stage 2)} are the 
predefined structures.}
% instructions for an explanation generator to generate the literal explanations of the reasoning process of a given LM.}
% Q, A, R, P and T are instructions for an explanation generator to generate the literal explanations of the reasoning process of a given LM. 
The generator outputs a natural language explanation in the form of a sentence or a paragraph. The details of our instruction are shown in Appendix (Figure~\ref{fig:instructions}).

\zc{\subsubsection{Enhancing Explanation Comprehension with LLM-based Debugging}}

The \textit{why-choose} and \textit{why-not-choose} explanations provide insights into the LM's reasoning behavior. However, for individuals without a strong background in AI, these explanations may still prove challenging to develop LMs. Inspired by the Transformer Debugger \citep{mossing2024tdb,bills2023language}, we include a debugging process named LM debugger to make explanations more accessible and understandable, regardless of the user's AI expertise. 
% an LM debugger tool associated with \methodname{} that leverages LLMs to make explanations more accessible and understandable, regardless of the user's AI expertise. 
The LM Debugger uses a LLM to simulate an ``prefect" LM (e.g., one with 100\% accuracy) and leverages explanations generated from \methodname{} to reconstruct the reasoning process of the target LM, which is the model under evaluation. It calculates the simulated reasoning score based on the differences between these simulations. We use a set of structured instructions to conduct the debugging process. The instructions are as follows: ``\textbf{System Prompt:} [MODEL\_NAME], \textbf{Explanation Content:} [TASK\_TYPE, $q$, $\mathcal{A}$, $y'$, $\mathcal{Q}$, $E$], \textbf{Evaluation Criteria:} \textit{[optional]}, \textbf{Advice Instruction:} \textit{[optional]}, \textbf{Debugging Instruction:} [MODEL\_NAME]". 
\textit{System Prompt}, \textit{Explanation Content}, \textit{Evaluation Criteria}, \textit{Advice Instruction}, and \textit{Debugging Instruction} are predefined instructions. 
The debugging process focuses on four aspects: faithfulness, completeness, minimality, and accuracy. By analyzing the explanations through these lenses, the LM Debugger offers users suggestions on potential areas for improvement within the LM and highlights any noteworthy observations. Our approach democratizes the understanding of LM reasoning behavior, empowering users from diverse backgrounds to engage with and benefit from the insights provided by the explanations. We provide a detailed description of the LM Debugger in the Appendix~\ref{app:lm_debugger}.

% \begin{table}[]
% \centering
% \resizebox{0.5\columnwidth}{!}{%
% \begin{tabular}{@{}c|l@{}}
% \toprule
% \begin{tabular}[c]{@{}c@{}}System\\  Prompt\end{tabular} & \begin{tabular}[c]{@{}l@{}}You're a professional researcher in NLP. \\ Write it step by step.\end{tabular}                                                 \\ \midrule
% Q             & Question content is...                                                                                         \\ \midrule
% A             & The predicted choice is...                                                                                     \\ \midrule
% R             & \begin{tabular}[c]{@{}l@{}}According to the model top reason-elements + \\ $\mathcal{K}$ + explain the model reasoning process with ``since...''\end{tabular}\\ \midrule
% P             & According to...                                                                                                \\ \midrule
% T             & \begin{tabular}[c]{@{}l@{}}Explain why the model doesn't choose other options \\ with ``The other potential choices...''\end{tabular}                          \\ \bottomrule
% \end{tabular}
% }
% \caption{The instructions for explanation generators. }
% \label{tab:instruction}
% \vspace{-10pt}
% \end{table}

\section{Experiments}

%\subsection{Dataset}
\subsection{Experiment Settings}
In our experiments, \zc{we evaluate the performance of the candidate approaches using two datasets: CommonsenseQA and OpenBookQA. CommonsenseQA consists of 12,247 five-way multiple-choice questions designed to test commonsense knowledge, while OpenBookQA comprises 5,957 four-way multiple-choice questions that assess a model's ability to reason with science knowledge. The key difference between these datasets lies in the nature of the knowledge being evaluated, making them complementary benchmarks for evaluating the reasoning capabilities of LMs across diverse domains.}
% we use the CommonsenseQA and OpenBookQA datasets to evaluate the performance of the candidate approaches. CommonsenseQA consists of 12,247 questions created by crowd-workers, which are designed to test commonsense knowledge through a 5-way multiple-choice QA task. OpenBookQA consists of 5,957 four-way multiple-choice questions designed to evaluate models' reasoning with elementary science knowledge.

%\subsection{Baselines}
Our evaluation can be divided into two parts. In the first part, we focus on \textbf{model performance}.
We compare~\methodname{} with two sets of baseline models on the CommonsenseQA and OpenBookQA. The first set comprises KG-augmented versions of RoBERTa-large. 
It includes the current SOTA LM+KG method on CommonsenseQA, MHGRN \citep{feng2020scalable}, KagNet~\citep{lin2019kagnet}, GconAttn~\citep{wang2019improving}, RGCN~\citep{schlichtkrull2018modeling}, RN~\citep{santoro2017simple},
QA-GNN \citep{yasunaga2021qagnn}, GreaseLM \citep{zhang2022greaselm}.
The second set consists of LLM Llama-2-7B~\citep{touvron2023llama-2}, which demonstrates the capabilities of LMs without interpretation. The LMs we used are from Huggingface\footnote{https://huggingface.co/}.

% consists of fine-tuned LM RoBERTa-large \citep{liu2019roberta} \jianda{and LLM Llama-2-7B~\citep{touvron2023llama-2}}, which demonstrate the capabilities of LMs without interpretation. 
% The second set of baseline models includes KG-augmented versions of RoBERTa-large, using ConceptNet as the source of common sense knowledge and following the approach in \citep{lin2019kagnet}. The third set of baseline models is the current SOTA commonsense reasoning method on CommonsenseQA, MHGRN \citep{feng2020scalable}, QA-GNN \citep{yasunaga2021qagnn}, GreaseLM \citep{zhang2022greaselm}. The LMs we used are from Huggingface\footnote{https://huggingface.co/}.

In the second part, we evaluate~\methodname{} on \textbf{explanation ability}. To establish a baseline for comparison, two prior works, namely PathReasoner \citep{ZHAN2022107612} and Explanations for CommonsenseQA (ECQA) \citep{aggarwal-etal-2021-explanations}, were employed as benchmarks. These works are recognized for providing natural and comprehensible explanations \zc{for model's reasoning behavior}.

%\subsection{Experiment Settings}
% \jianda{We train two variants of \methodname{},
% % our method, \methodname{}(RoBERTa-large) and \methodname{}(Llama-2-7B),
% which comprises with RoBERTa-large and Llama-2-7B, respectively, as the LM.} 

We train two variants of \methodname{}, each utilizing a different LM: RoBERTa-large and Llama-2-7B, respectively. We train the models on a single NVIDIA A100 GPU. Further details regarding the model hyperparameters, training settings, and knowledge graph preprocessing steps can be found in Appendix \ref{app:experimental_settings}.

\subsection{Experimental Results}

We present our experimental results in Table~\ref{tab:acc} and Table~\ref{tab:bookacc}, where the accuracy of~\methodname{} is evaluated on the CommonsenseQA and OpenBookQA datasets. Our empirical findings indicate that~\methodname{} leads to consistent improvements in performance compared to existing baseline methods on both datasets. Specifically, the test performance on CommonsenseQA is improved by 4.71\% over the prior best LM+KG method, GreaseLM, 5.35\% over the included KG augmented LMs, and 7.12\% over fine-tuned LMs. 
The test performance achieves comparable results to the prior best LM+KG method, GreaseLM, on OpenBookQA. 
However, our proposed~\methodname{} utilizing LLM Llama-2 significantly outperforms baseline LM+KG method by 8.6\%. 
It is worth noting that LLM \mbox{Llama-2} is trained with a huge amount of data, so that 
finetuning LLM Llama-2 without KG is able to achieve comparable results to ~\methodname{}.

\noindent
\begin{minipage}[htbp]{0.48\textwidth}
  \centering
  \resizebox{\columnwidth}{!}{%
  \begin{tabular}{lrr}
    \toprule
    \textbf{Method}                             & \textbf{IHdev-Acc.} & \textbf{IHtest-Acc.} \\
    \midrule
    \textbf{Baselines} \citep{feng2020scalable} &                     &                      \\
    MHGRN (2020)                                & 73.69\%             & 71.08\%              \\
    KagNet (2019)                               & 73.47\%             & 69.01\%              \\
    GconAttn (2019)                             & 72.61\%             & 68.59\%              \\
    RGCN (2018)                                 & 72.69\%             & 68.41\%              \\
    RN (2017)                                   & 74.57\%             & 69.08\%              \\
    \midrule
    \textbf{Baselines} (our implementation)     &                     &                      \\
    GreaseLM (2022)                             & 76.17\%             & 72.60\%              \\
    QA-GNN (2021)                               & 74.94\%             & 72.36\%              \\
     Llama-2-7B (w/o KG) (2023)              & 81.49\%    & 78.24\%     \\
    \midrule
   ~\methodname{} (RoBERTa-large)            & \textbf{77.97\%}    & \textbf{77.31\%}     \\ 
   ~\methodname{} (Llama-2-7B)               & \textbf{82.88\%}    & \textbf{77.36\%}     \\
    \bottomrule
  \end{tabular}
  }
  \captionof{table}{Comparative performance of \methodname{} on CommonsenseQA In-House Split: Our model surpasses all baselines, achieving accuracies of 77.97\% and 77.31\% with RoBERTa-large, and 82.88\% and 77.36\% with Llama-2-7B on IHdev and IHtest, respectively. While the LMExplainer (Llama-2-7B) closely matches the performance of Llama-2-7B, it offers the benefit of explainability.}
  \label{tab:acc}

\end{minipage}\hfill
\begin{minipage}[htbp]{0.48\textwidth}

  \centering
  \resizebox{\columnwidth}{!}{%
  \begin{tabular}{lrr}
    \toprule
    \textbf{Method}                             & \textbf{Dev-Acc.} & \textbf{Test-Acc.} \\
    \midrule
    \textbf{Baselines} \citep{feng2020scalable} &                   &                    \\
    MHGRN (2020)                                & 68.10\%           & 66.85\%            \\
    GconAttn (2019)                             & 64.30\%           & 61.90\%            \\
    RGCN (2018)                                 & 64.65\%           & 62.45\%            \\
    RN (2017)                                   & 67.00\%           & 65.20\%            \\
    \midrule
    \textbf{Baselines} (our implementation)     &                   &                    \\
    GreaseLM (2022)                             & 71.80\%           & 70.80\%   \\
    QA-GNN (2021)                               & 63.00\%           & 59.80\%            \\
    Llama-2-7B (w/o KG) (2023)            & 80.60\%    & 78.40\%     \\
    \midrule
   ~\methodname{} (RoBERTa-large)                          &\textbf{ 69.20\%}           & \textbf{68.00\%}           \\
   ~\methodname{} (Llama-2-7B)                          & \textbf{80.80\%}           & \textbf{79.40\%}            \\   
    \bottomrule
  \end{tabular}
  }
  \captionof{table}{Performance Comparison on OpenBookQA: 
  ~\methodname{} demonstrates competitive results against various baselines, closely matching the best. Notably, while GreaseLM is optimized for accuracy in QA tasks, \methodname{} focuses on explaining the reasoning process behind its answers. \methodname{} + Llama-2-7B achieves the best performance, combining high accuracy with the added value of explainability. }
  \label{tab:bookacc}
  \vspace{8pt}
\end{minipage}

Beyond achieving high accuracy, \methodname{} also provides transparency in reasoning, enhancing human understanding of the LM's decision-making behavior.
% It is worth noting that GreaseLM is specifically designed to improve accuracy for QA tasks, while our~\methodname{} focuses on providing explanations for the reasoning process. Despite this difference in focus, our~\methodname{} not only offers insight into the underlying reasoning but also demonstrates an improvement in performance. This finding highlights the potential benefits of incorporating explainability into the model design, as it may lead to enhanced performance in addition to fostering a better understanding of the decision-making process.
This finding highlights the potential benefits of incorporating interpretability components that narrow the reasoning space into the model architecture. By focusing on the most relevant elements, these components not only facilitate a better understanding of the decision-making behavior but also contribute to enhance model performance.
To more thoroughly understand the influence of various components of \methodname{} on its overall performance, we have conducted a ablation study in Appendix \ref{sec:ablation}.

\vspace{-1mm}
\subsection{Explanation Results}

\definecolor{green}{HTML}{B7C47D}
\definecolor{red}{HTML}{D46155}

\begin{figure}[htb]
\centering

\begin{tcolorbox}[colback=softGray, colframe=deepBlue, title=\ding{52} Explanation Example of Correct Prediction, 
fonttitle=\fontsize{8pt}{1pt}\selectfont, fontupper=\fontsize{5pt}{1pt}\selectfont,]
\textbf{Input:} 
The question is ``What is someone doing if he or she is sitting quietly and his or her eyes are moving?''\\
The Answer Options are:
A. Reading, B. Meditate, C. Fall Asleep, D. Bunk, E. Think

\textbf{Output (Model Prediction):} “A. Reading”

\textbf{Top-ranked reason-elements:}
1. Quiet chattering mind, 2. Not making sound, 3. Mind focuses, 4. Glasses for people with poor eyesight, 5. War
\vspace{2mm} 
\hrule 
\vspace{2mm} 
\textbf{Stage 1 - Explanation (why-choose):}
The model predicts "Reading" (A) as the activity someone is engaged in when sitting quietly with moving eyes, based on the key reason-elements. Here's how the reason-elements support this conclusion:\\
- \highlight{\textbf{Quiet Chattering Mind \& Not Making Sound:}} Indicate silent, internal focus, typical of reading, fitting the description of someone sitting quietly.\\
- \highlight{\textbf{Mind Focuses:}} Suggests active engagement with content, aligning with reading.\\
- \highlight{\textbf{Glasses for Poor Eyesight:}} Implies a visual task, supporting reading despite being an indirect clue.\\
% - \highlight{\textbf{War:}} Irrelevant to the decision-making process.

Excluding outliers, these elements collectively signal silent, visually focused activity, characteristic of reading.

\textbf{Stage 2 - Explanation (why-not-choose):}
The model's reasoning for making other options less likely is as follows:\\
- \highlight{\textbf{Meditate (B):}} Involves closed or unfocused eyes, unlike reading's active eye movement.\\
- \highlight{\textbf{Fall Asleep (C):}} Entails cessation of eye movement, not alignment with reading.\\
- \highlight{\textbf{Bunk (D):}} Lacks specific association with quiet, focused eye movement.\\
- \highlight{\textbf{Think (E):}} Doesn't necessitate visual engagement, different from reading.

Thus, only reading aligns with the criteria of silent, focused activity involving active visual engagement.

\vspace{2mm} 
\hrule 
\vspace{2mm} 
\textbf{LM Debugger:}

- \highlight{\textbf{Faithfulness:}} 4/5 - Well-aligned with the LM's data-driven reasoning.

- \highlight{\textbf{Completeness:}} 4/5 - Captures key reasons for "Reading" but may miss minor factors distinguishing it from similar activities.

- \highlight{\textbf{Minimality:}} 4/5 - Includes mostly relevant reasoning but is diluted by the unrelated "War."

- \highlight{\textbf{Accuracy:}} 4/5 - Accurately identifies "Reading" as the most likely activity, with minor issues in relevancy and nuanced differentiation.

- \highlight{\textbf{Advice for Improvement:}} Enhance the LM by broadening training data to cover more \textbf{physical activities} for accuracy, and fine-tuning with \textbf{detailed activity datasets} for sharper recognition.
\end{tcolorbox}
\caption{Example of \methodname{}'s explanation for a correct LM prediction, with high scores and potential enhancement suggestions from the LM Debugger on key dimensions. }
\vspace{-5mm}

\label{fig:correct_expl}

\end{figure}

\begin{wraptable}{r}{0.6\textwidth} 
  \vspace{-14mm}
  \centering
  \resizebox{0.58\textwidth}{!}{%
    \begin{tabular}{@{}c|l@{}}
      \toprule
      \textbf{\makecell{PathReasoner \\ \citep{ZHAN2022107612}}} & \begin{tabular}[c]{@{}l@{}} \textbf{Output (Model Prediction):} ``A. Reading'' \\\textbf{Explanation:} \\
      quietly [related to] quiet [at location] a library [used for] reading \\\end{tabular} \\ \midrule
      \textbf{\makecell{ECQA \\\citep{aggarwal-etal-2021-explanations}}} & \begin{tabular}[c]{@{}l@{}} \textbf{Output (Model Prediction):} ``A. Reading'' \\\textbf{Explanation:} \\ While meditating and sleeping, eyes don't move, eyes are closed. \end{tabular} \\ \bottomrule
    \end{tabular}%
  }
  \caption{Examples from PathReasoner and ECQA illustrate the LM's reasoning, offering insights yet lacking in clarity and faithfulness. See Appendix (Table~\ref{app:example}) for a detailed comparison.}
  \label{tab:explanation}
  % \vspace{-2mm}
\end{wraptable}

Our explanation results are in Figure~\ref{fig:correct_expl} and Table~\ref{tab:explanation}. 
The LM $f_{LM}$ used in our explanation is RoBERTa-large, paired with GPT-4-turbo \citep{OpenAI2023GPT4TR} as the explanation generator and LM Debugger. Note that  $f_{LM}$ serves as a representative example and can be replaced with other LMs as required. To further demonstrate the effectiveness of our approach, we compare it with two SOTA methods, PathReasoner \citep{zhan2022pathreasoner} and ECQA \citep{aggarwal-etal-2021-explanations}. PathReasoner uses structured information to explain the reasoning path, while ECQA first is created by human-annotated explanations and then leverages a generation model to organize the final explanation. 
% In Table~\ref{tab: explanation}, we present the inputs and results of our approach, which include ranked \emph{reason-elements} and explanations of the reasoning process. These examples highlight the ability of~\methodname{} in generating comprehensive and interpretable explanations for the LMs. Complete results are shown in Appendix \ref{sec:appendix-example}.

% In comparison to PathReasoner explanations, which only provide structured reasoning paths that are non-informative and require manual selection of a specific path, our proposed approach not only offers a complete reasoning path but also provides a justification for the predicted answer. 
As illustrated in Table~\ref{tab:explanation} and Table~\ref{app:example} (complete results), PathReasoner presents four reasoning paths, including redundant paths, making it difficult to identify the faithful reasoning path. % consists of human-annotated explanations that provide highly accurate descriptions of the reasoning process. However, its explanations are 
While ECQA simply combines the positive and negative examples, which fails to explain the actual decision-making process of the LM. 
% In contrast, our explanations are not a mere combination of sentences but are inferred and logically derived. \methodname{} provides a more comprehensive and accurate depiction of the LM's actual reasoning behavior and improves the overall interpretability. 
% and usefulness of the generated explanations. 
% In addition, the \emph{why-not-choose} explanation explains why the model does not choose other answers, which gives people a better understanding of the model’s reasoning behavior and increases the transparency of the LM.
Compared to other explanation methods, \methodname{} provides a more comprehensive and structured explanation, covering both the chosen answer and the reasons for not selecting other options. The LM Debugger's evaluation and advice for improvement are unique features of \methodname{}, offering insights into the explanation's quality and potential areas for enhancing the LM's performance.
We provide a comprehensive case study in Appendix \ref{app:case}.

% In contrast, our method provides a clear and concise natural language explanation for the chosen answer (\emph{why-choose} explanation), which greatly enhances the understandability and smoothness of the explanation.

% The ECQA consists of human-annotated explanations that provide highly accurate descriptions of the reasoning process. However, as shown in Table~\ref{tab:explanation} and Table~\ref{app:example}, its explanations are simply a combination of positive and negative examples provided by humans, which fails to illustrate the actual decision-making process of the model.
% While this approach can generate high-quality explanations from a human perspective, it fails to illustrate the actual decision-making process of the model. 
% In contrast, our explanations are not a mere combination of sentences but are inferred and logically derived.~\methodname{} provides a more comprehensive and accurate depiction of the reasoning process and improves the overall interpretability and usefulness of the generated explanations. In addition, the \emph{why-not-choose} explanation explains why the model does not choose other answers, which gives people a better understanding of the model’s reasoning process and increases the transparency of the model.
% For an in-depth understanding, a detailed case study is available in Appendix \ref{sec:case}.

% \vspace{-1mm}
\subsection{Evaluation of Explanations}
% \vspace{-1mm}
We evaluate the \methodname{} generated explanations from two perspectives: simulated reasoning and user perception. 
% The simulated reasoning perspective evaluates the explanations' alignment with the reasoning process of an ``prefect'' LM, using the LM Debugger. 
% This evaluation helps users understand the reasoning behind the explanations, assess the LM's behavior, and identify the potential limitations. 
% This evaluation helps identify areas where the actual LM's capabilities can be improved to more closely mirror the expected behavior of the simulated 'ideal' LM. 
% The user perception perspective focuses on the effectiveness of \methodname{} in helping users understand the LM's reasoning process. 
% We evaluate the \methodname{} generated explanations from two aspects: quantitative and qualitative. The quantitative aspect assesses the quality of explanations using the LM Debugger, which serves as an evaluator of explanation quality and helps users interpret and trust the explanations provided by \methodname{}. The qualitative aspect focuses on the effectiveness of \methodname{} in helping users understand the LM's reasoning process. 
We randomly select 20 QA pairs from CommonsenseQA dataset, using RoBERTa-large as the LM ($f_{LM}$, acc=77.97\%) and GPT-3.5-turbo as the explanation generator and GPT-4-turbo as LM Debugger.

\begin{wrapfigure}{r}{0.5\textwidth}
    \centering
    \vspace{-12mm}
    \includegraphics[width=1\linewidth]{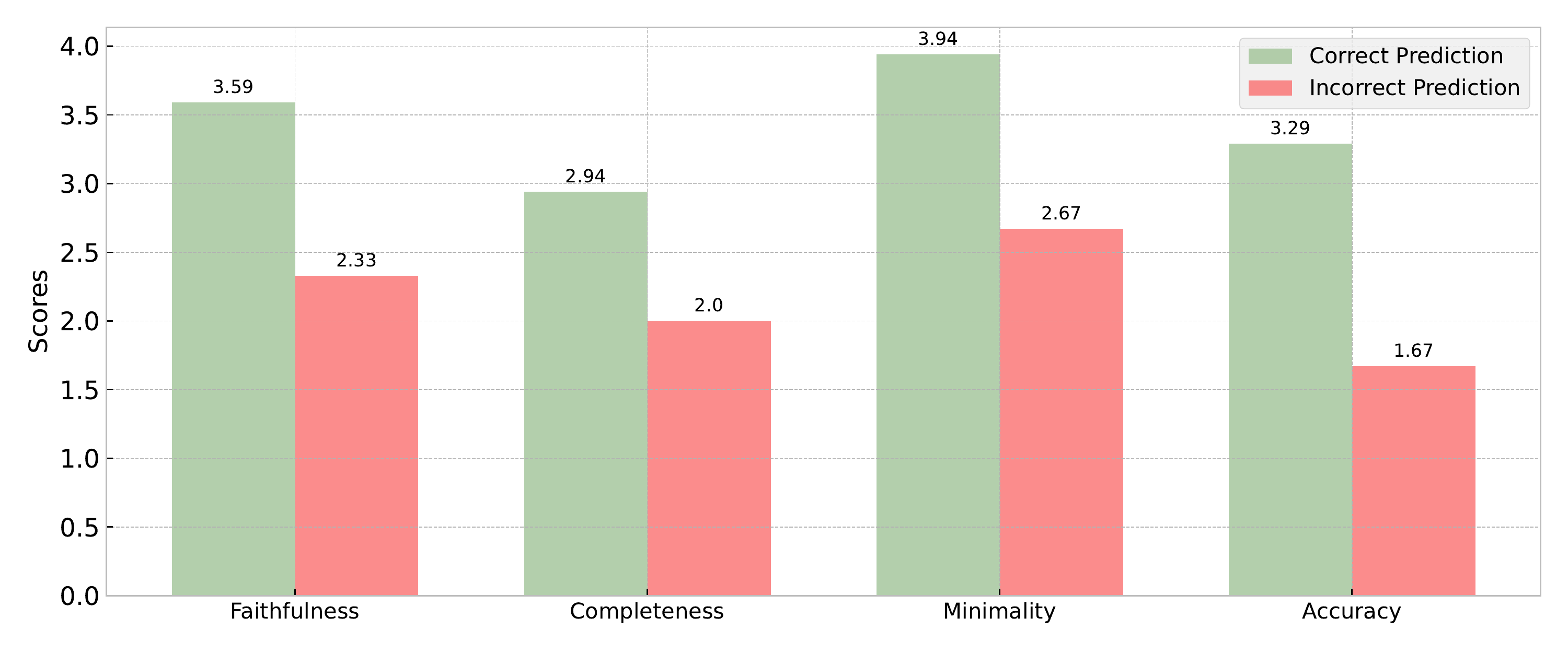}
    \caption{Simulated reasoning scores split by LM's prediction correctness.}
    \label{fig:explanation_quality}
    \vspace{-5mm}
\end{wrapfigure}
\zc{\subsubsection{Simulated Reasoning Perspective}}
The simulated reasoning perspective evaluates the explanations' alignment with the reasoning process of a ``prefect'' LM, using the LM Debugger. 
This evaluation helps users understand the reasoning behind the explanations, assess the LM's behavior, and identify the potential limitations. 

We evaluate the explanations across four key dimensions from a debugging perspective, following the evaluation metrics of \citep{bills2023language}: faithfulness, completeness, minimality, and accuracy. Each dimension is scored on a scale from 1 to 5, utilizing strict scoring criteria, where higher scores indicate better performance. The details of the scoring criteria and effectiveness evaluation of LM Debugger are shown in Appendix \ref{app:lm_debugger}.

% Figure~\ref{fig:} presents the effectiveness of the LM Debugger.
% We set a threshold score of 3. Explanations with scores equal to or above this threshold are considered satisfactory. 

% The results demonstrate the high effectiveness of the LM Debugger in accurately identifying the quality of explanations based on the model's prediction correctness. For both correct and incorrect predictions, the LM Debugger achieves 100\% accuracy in assessing faithfulness and accuracy. It also performs well in identifying completeness, with 88.24% accuracy for correct predictions and 100% for incorrect predictions.

% \begin{table}[htb]
% \centering
% \begin{tabular}{lcc}
% \hline
% \textbf{Dimension} & \textbf{Correct Prediction} & \textbf{Incorrect Prediction} \\
% \hline
% Faithfulness & 100\% & 100\% \\
% Accuracy & 100\% & 100\% \\
% \hline
% \end{tabular}
% \caption{Percentage of cases where the LM Debugger correctly identifies the reliability of the model's predictions based on explanation quality.}
% \label{tab:explanation_quality}
% \end{table}

% ---
% \begin{table}[htb]
% \centering
% \begin{tabular}{lcccc}
% \hline
% \textbf{Dimension} & \textbf{Correct Prediction} & \textbf{Incorrect Prediction} \\
% \hline
% Faithfulness & 3.59 & 2.33 \\
% Completeness & 2.94 & 2.00 \\
% Minimality & 3.94 & 2.67 \\
% Accuracy & 3.29 & 1.67 \\
% \hline
% \end{tabular}
% \caption{Explanation quality scores split by prediction correctness.}
% \label{tab:explanation_quality_correctness}
% \end{table}

The simulated reasoning scores are presented in Figure \ref{fig:explanation_quality}, showing a striking difference in scores between correct and incorrect predictions. This allows users to easily distinguish between reliable and unreliable LM results. We provide a comparative case study in Appendix \ref{app:case}, showcasing how explanations differ when the LM's prediction is correct or incorrect, highlighting \methodname{}'s effectiveness in generating informative explanations.

% \subsubsection{Evaluation of Explanation}
\zc{\subsubsection{User Perception Perspective}}
\begin{table*}[ht]
\resizebox{\textwidth}{!}{%
\footnotesize
\begin{tabular}{@{}l|c|ccccccc@{}}
\toprule
{} & {} & \multicolumn{7}{c}{\textbf{Evaluation Metrics}} \\
\cline{3-9}
\multirow{-2}{*}{\textbf{Evaluator}} & \multirow{-2}{*}{\textbf{Strong AI Understanding?}} & \textbf{Overall Quality} & \textbf{Understandability} & \textbf{Trustworthiness} & \textbf{Satisfaction} & \textbf{Sufficiency of Detail} & \textbf{Completeness} & \textbf{Accuracy} \\ \midrule
Human Experts & \textcolor{green}{\ding{52}} & 0.91 & 0.97 & 0.95 & 0.89 & 0.98 & 0.97 & 0.93 \\ 
Crowdsourcing & \textcolor{red}{\ding{56}} & 0.85 & 0.89 & 0.86 & 0.80 & 0.83 & 0.81 & 0.85 \\ 
GPT-3.5 & \textcolor{green}{\ding{52}} & 0.98 & 0.98 & 0.98 & 0.98 & 0.98 & 0.98 & 0.98 \\
GPT-4 & \textcolor{green}{\ding{52}} & 0.90 & 0.93 & 0.87 & 0.87 & 0.88 & 0.87 & 0.88 \\ \bottomrule
\end{tabular}

}
\caption{Evaluation by automated evaluator GPT-3.5-turbo, GPT-4-turbo, human experts, and crowdsourcing on 7 evaluation metrics. \zc{Our explanations are accessible to all regardless of AI knowledge.}}
\vspace{-10pt}
\label{tab:simulator_score}
\end{table*}

The user perception perspective focuses on the effectiveness of \methodname{} in helping users understand the LM's reasoning process. Our user perception evaluation includes three approaches: human expert review, crowdsourcing, and automated methods. Our expert panel consists of individuals with graduate-level education, taught in English, and a minimum of three years of research experience in NLP. We also hire 50 general users through the crowdsourcing platform Prolific~\footnote{https://www.prolific.com}, ensuring a gender-balanced participant pool of native English speakers, all possessing at least a high school education. For automated evaluation, we utilize GPT-3.5-turbo and GPT-4-turbo to further validate the explanations. The evaluation follows the methodology in \citep{hoffman2018metrics} and involves seven evaluative dimensions: overall quality, clarity, credibility, satisfaction, detail adequacy, completeness, and accuracy. Participants rate these aspects using a three-point Likert scale, and scores are normalized to a range $[0,1]$, with higher scores indicating better quality. 

The scores are shown in Table \ref{tab:simulator_score}. Human experts highly commend the Understandability, Trustworthiness, and Completeness (above 0.95) of our explanations. They acknowledge our adeptness in producing comprehensive and reliable explanations. The crowdsourcing results are slightly lower across all metrics. This outcome potentially mirrors the diverse and less specialized viewpoints of the general public. Overall, the general users are able to understand how LMs reason through our explanations. Automated evaluators GPT-3.5-turbo and GPT-4-turbo provide evaluations of our explanations' quality that closely match human expert assessments, demonstrating consistency across key metrics.
GPT-3.5-turbo agrees with our strong performance in Overall Quality, Understandability, and Accuracy, with each scoring 0.98. Similarly, GPT-4-turbo gives a comparable evaluation, with its highest score in Understandability (0.93). 

% The notably lower scores in ``Irrelevance'' indicate incorrect inferences result in irrelevant information in our explanations. This issue, easily identified by evaluators, highlights a potential area for future human-centered explanations.

These results highlight the high quality of our explanations. Notably, even participants without a strong AI background provided high scores, indicating the accessibility and effectiveness of \methodname{}'s explanations for a wide audience. The consistency across key metrics emphasizes the effectiveness and reliability of the explanations generated by \methodname{}. The details of the automated evaluation process and questionnaires are outlined in Appendix \ref{appendix:auto-eval}.

\section{Conclusions}
In this paper, we propose~\methodname{}, a novel model that incorporates an interpretation module to enhance the performance of LMs while also providing grounded and trustworthy explanations of the LM's reasoning behavior.
Our explanation results are presented in a logical and comprehensive manner, making it easier for humans to understand the model's reasoning in natural language. 
\zc{The debugging process helps users to refine LMs for improved performance, marking a step towards democratizing AI.}
Our experimental results demonstrate superior performance compared to prior SOTA works across standard datasets in the commonsense domain. Our analysis shows that~\methodname{} not only improves the model's performance but also provides humans with a better understanding of the model.

% \paragraph{Limitation.}
% Although datasets like CommonsenseQA and OpenBookQA cover various topics, they don't fully represent the diverse domains where LMs are applied. Future work will evaluate \methodname{}'s effectiveness in specialized fields such as legal and medical.

\section{Ethics Statement}
The primary ethical concern in our work relates to the use of LLMs for explanation generation.
Specifically, if the explanation generator is of low quality or deemed unsafe, it presents a significant risk. This could adversely affect the integrity and reliability of the content and style of the explanations. We acknowledge the importance of ensuring the quality and safety of the explanation generator to maintain ethical standards in our outputs and to prevent the dissemination of potentially harmful or misleading information.

\bibliography{conference}
\bibliographystyle{conference}

\newpage
\appendix
\section{Algorithms}

We present the details of two fundamental algorithms of \methodname{}. Algorithm ``Construct Element-graph'' shows  the procedure for generating a pruned element-graph \(G_e\) from the input content \(z\). It utilizes LM and ConceptNet to extract relevant nodes, evaluates them to compute a pruning score, and selects the top \(K\) nodes to form \(G_e\).

Algorithm ``Element-graph Interpretation'' processes the pruned element-graph \(G_e\) to identify essential ``Reason-elements'' that enhancing the understanding of \(z\). It uses a graph attention network, which utilizes attention to analyze node interactions, updates node features, and calculates the probability of selecting specific outcomes. This process results in the identification and ranking of nodes to highlight important ``Reason-elements.''

\begin{algorithm} [H]
\DontPrintSemicolon
\KwData{Input content $z$}
\KwResult{Pruned element-graph $G_e$}
\caption{Construct Element-graph}
\label{algo:subgraph}

\Begin{
    $G_k \gets \text{ExtractFromConceptNet}(z)$ \\ \tcp*[r]{Extract the $L$-hop neighbor from ConceptNet}
    \For{each node $v_e$ in $G_k$}{
        % $v_{\text{embedding}} \gets f_{\text{enc}}(v)$ \tcp*[r]{Embedding function of $f_{LM}$} 
     %   $v_{\text{embedding}} \gets z_{emb} \mathbin\Vert v_{emb}$ \tcp*[r]{Concat text embeddings}
        $v_{\text{score}} \gets f_{prob}(z_{emb},v_{emb})$ \tcp*[r]{Compute score for pruning}
    }
    $G_e \gets \text{SelectTopK}(G_k)$ \tcp*[r]{Prune based on top $K$ scores}
    \Return{$G_e$}\;
}
\end{algorithm}

\begin{algorithm}[H]
\DontPrintSemicolon
\KwData{Element-graph \( G_e \)  containing node type embedding \(u_i\) and relation embedding \(r_{ij}\), input \(z\).}
\KwResult{\emph{Reason-elements}}
\caption{Element-graph Interpretation}
\label{algo:element}

\Begin{
        \For{each attention layer \( k \) in graph reasoning network}{
        \For{each node \( v_i \) in \( G_e \)}{
      \(\alpha_{ij}  \gets  \frac{\exp(a(h_{ki},h_{kj},u_i,r_{ij}))}{\sum \limits_{v_l \in \mathcal{N}_i } \exp(a(h_{ki},h_{kl}))}\) \tcp*[r]{Compute attention coefficient \(\alpha_{ij}\)}
     \(  h_{k+1,i}  \gets  f_{\delta}\left(\sum\limits_{v_j\in \mathcal{N}_i}   \alpha_{ij} m(h_{kj},u_i,r_{ij}) \right) + h_{ki} \) \tcp*[r]{Update node feature}
      }
    }
  %  \( \Lambda \gets \text{AttentionPooling}(G_e) \) \tcp*[r]{Attention-based pooling}
   % \( \mathbb{H}^{itp} \gets h_{v_i}^k \parallel \Lambda,\; \mathbb{H}^{LM} \gets f_{LM}(z) \) \tcp*[r]{Forming \(\mathbb{H}^{itp}\) and \(\mathbb{H}^{LM}\)}
  %  \( P(a|q) \propto \exp(\text{MLP}(\mathbb{H}^{LM}, \mathbb{H}^{itp})) \) \tcp*[r]{Probability of choosing an answer}
 \(\mathbb{H}^{LM} \gets f_{LM}(z) \) \tcp*[r]{Forming  \(\mathbb{H}^{LM}\)}
 \( P(y'|q) \propto \exp(\text{MLP}(\mathbb{H}^{LM}, \boldsymbol{h}_K,\boldsymbol{\alpha}_K)) \) \tcp*[r]{Probability of choosing an answer}
 \( \text{ReasonElements} \gets \text{RankNode}(G_e, \boldsymbol{\alpha}_K)) \) \tcp*[r]{Rank nodes based on the attentions}

    \Return{ReasonElements}
}
\end{algorithm}

\section{Experimental Settings}\label{app:experimental_settings}

We set our GNN module to have 200 dimensions and 5 layers, where a dropout rate of 0.2 is applied to each layer. We train the model on a single NVIDIA A100 GPU with a batch size of 64. The learning rates for the language model and the GNN module are set to $1e-5$ and $1e-3$, respectively. 
We opt for the RAdam optimizer for RoBERTa-large, while employing AdamW for Llama-2-7B.
These settings are adopted in the first part of the evaluation to investigate the performance of the GNN module.

We employ ConceptNet \citep{speer2017conceptnet} as our external knowledge source for CommonsenseQA and OpenBookQA. ConceptNet contains a vast amount of information with 799,273 nodes and 2,487,810 edges, which provides a valuable resource for improving the accuracy of QA systems. We extract the $G_k$ with a hop size of 2, and subsequently prune the obtained graph to retain only the top 200 nodes.

\section{Instruction for Explanation Generation}

\begin{figure}[h!]
\centering
% \resizebox{0.95\textwidth}{!}{%
\begin{stage1}
\textbf{Basis:} Given a LM augmented with a graph attention network to extract key reasoning elements for decision-making. 
The task is \highlight{[TASK\_TYPE]}.

\textbf{Input:} The question is: \highlight{[$q$]}. The Answer Options are: \highlight{[$\mathcal{A}$]}

\textbf{Output:} The model predicted choice \highlight{[$y'$]}. Based on the Ranked Reason-elements: \highlight{[$\mathcal{Q}$]}

\textbf{Explanation (Stage 1):} 
Explain the LM's reasoning process for selecting \highlight{[$y'$]} over the other options. Provide concise explanations for why each reason-element supports \highlight{[$y'$]} as the predicted choice. Focus on the LM's behavior and the significance of the Ranked Reason-elements. Your response should be short and concise.

\textbf{Explanation (Stage 2):} Based on the \highlight{[$E_0$]}, explain why this LM makes the other options less likely \highlight{[$\mathcal{A}\setminus\{y'\}$]}. Your response should be short and concise.
\end{stage1}
% }
\caption{The instructions for explanation generators.}
\label{fig:instructions}
\end{figure}

\section{LM Debugger} \label{app:lm_debugger}

The LM Debugger is an important component of the \methodname{}, designed to simulate the reasoning process of a ``perfect'' LM using the generated explanations. By treating the explanations as a representation of the LM's decision-making process, the debugger evaluates the quality of these explanations across four key dimensions: faithfulness, completeness, minimality, and accuracy. Additionally, it offers advice for improving the actual LM's performance based on the simulated reasoning process.

\subsection{Evaluation Dimensions}

\paragraph{Faithfulness} The LM Debugger evaluates whether the explanation faithfully represents the simulated LM's reasoning process. It checks if the provided rationale aligns with the data-driven and algorithmic processes an simulated LM would use to arrive at its conclusions. This evaluation ensures that the explanation accurately reflects the expected decision-making process of a simulated LM, serving as a benchmark for the actual LM's performance.

\paragraph{Completeness} The LM debugger determines whether the explanation fully captures the breadth of data that an simulated LM would leverage to make a decision. It ensures that no significant computational strategies or data insights that a simulated LM would rely on are omitted. A complete explanation should provide a comprehensive understanding of the simulated LM's decision-making process, including all relevant factors and considerations.

\paragraph{Minimality} The LM Debugger verifies that the explanation includes only the essential computational processes or data insights that an simulated LM would utilize, without unnecessary elaboration or speculative reasoning beyond the expected operational framework. This evaluation helps to maintain the reasoning's conciseness and relevance, focusing on the key factors that directly influence the simulated LM's decision-making process.

\paragraph{Accuracy} The LM Debugger acts as a simulated "perfect" LM, using the generated explanations to assess how well they align with the reasoning process and capabilities of an simulated LM. It evaluates the accuracy of the explanations against the simulated LM's standard. 

\subsection{Advice for Improvement}

Based on the evaluation results, the LM Debugger suggests improvements focusing on enhancing the actual LM's training data diversity, algorithmic transparency, or decision-making accuracy to better align with the simulated ``prefect'' LM. These suggestions aim to mitigate biases and increase the LM's reliability and trustworthiness. 

\subsection{Instruction for LM Debugger}

\begin{figure}[h]
\centering
\begin{tcolorbox}[colback=softGray, colframe=deepBlue, title=Instruction for LM Debugger, fontupper=\fontsize{8pt}{1pt}\selectfont,]

\textbf{System Prompt:} Evaluate the explanation provided for the LM‘s decision-making process in answering the given question. Assess the explanation across four dimensions: faithfulness, completeness, minimality, and accuracy. Assume the role of an LM debugger with expertise in the inner workings of \highlight{[MODEL\_NAME]} and a 100\% accurate \highlight{[MODEL\_NAME]}.

\textbf{Explanation Content:} Given a LM augmented with a graph attention network to extract key reasoning elements for decision-making. The task is \highlight{[TASK\_TYPE]}. The question is \highlight{[$q$]}. The Answer Options are: \highlight{[$\mathcal{A}$]}. The LM's prediction is \highlight{[$y'$]}. Top-ranked reason-elements are \highlight{[$\mathcal{Q}$]}. Explanation:
\highlight{[$E$]}. 

\textbf{Evaluation Criteria:} 

- Faithfulness: Assess if the explanation truly represents the LM's underlying computational and statistical mechanisms. Check if the rationale provided mirrors the data-driven and algorithmic processes the LM uses to arrive at its conclusions.
% Does the explanation accurately reflect the model's reasoning process? Consider if the explanation could independently arrive at the same conclusion as the whole model.

- Completeness: Determine whether the explanation fully captures the breadth of data and algorithms the LM leverages to make a decision. Ensure no significant computational strategies or data insights that the LM relies on are omitted.
% Evaluate whether the explanation encompasses all the elements used by the model to perform the task. Are any critical reasoning elements missing from the explanation?

- Minimality: Verify that the explanation includes only the essential computational processes and data insights the LM utilizes, without unnecessary elaboration or speculative reasoning beyond the LM's actual operational framework.
% Assess if the explanation contains only the necessary elements relevant to the task. Identify any nodes or parts of the explanation that are irrelevant or redundant in justifying the model's decision.

- Accuracy: Evaluate the precision with which the explanation reflects the LM's true capabilities and decision-making process, taking into account the LM's design, training data, and algorithmic functions.
% Based on the results of faithfulness, completeness, and minimality, assess the accuracy of the answer.

\textbf{Advice Instruction:} 
Suggest improvements focusing on enhancing the LM's training data diversity, algorithmic transparency, or decision-making accuracy, to mitigate biases and increase the model's reliability and trustworthiness in varied contexts.

\textbf{Debugging Instruction:}
Please provide a score from 1 to 5 for each dimension, with 1 being the lowest (poor, equivalent to \highlight{[MODEL\_NAME]} having 0\% accuracy) and 5 being the highest (excellent, equivalent to \highlight{[MODEL\_NAME]} having 100\% accuracy). Highlight specific areas where the explanation aligns well or falls short of the evaluation criteria. Your response should be short and concise.

\end{tcolorbox}
\caption{The instructions for LM Debugger.}
\end{figure}

\subsection{Evaluation of Effectiveness}

\begin{figure}[htbp]
    \centering
        \includegraphics[width=0.4\textwidth]{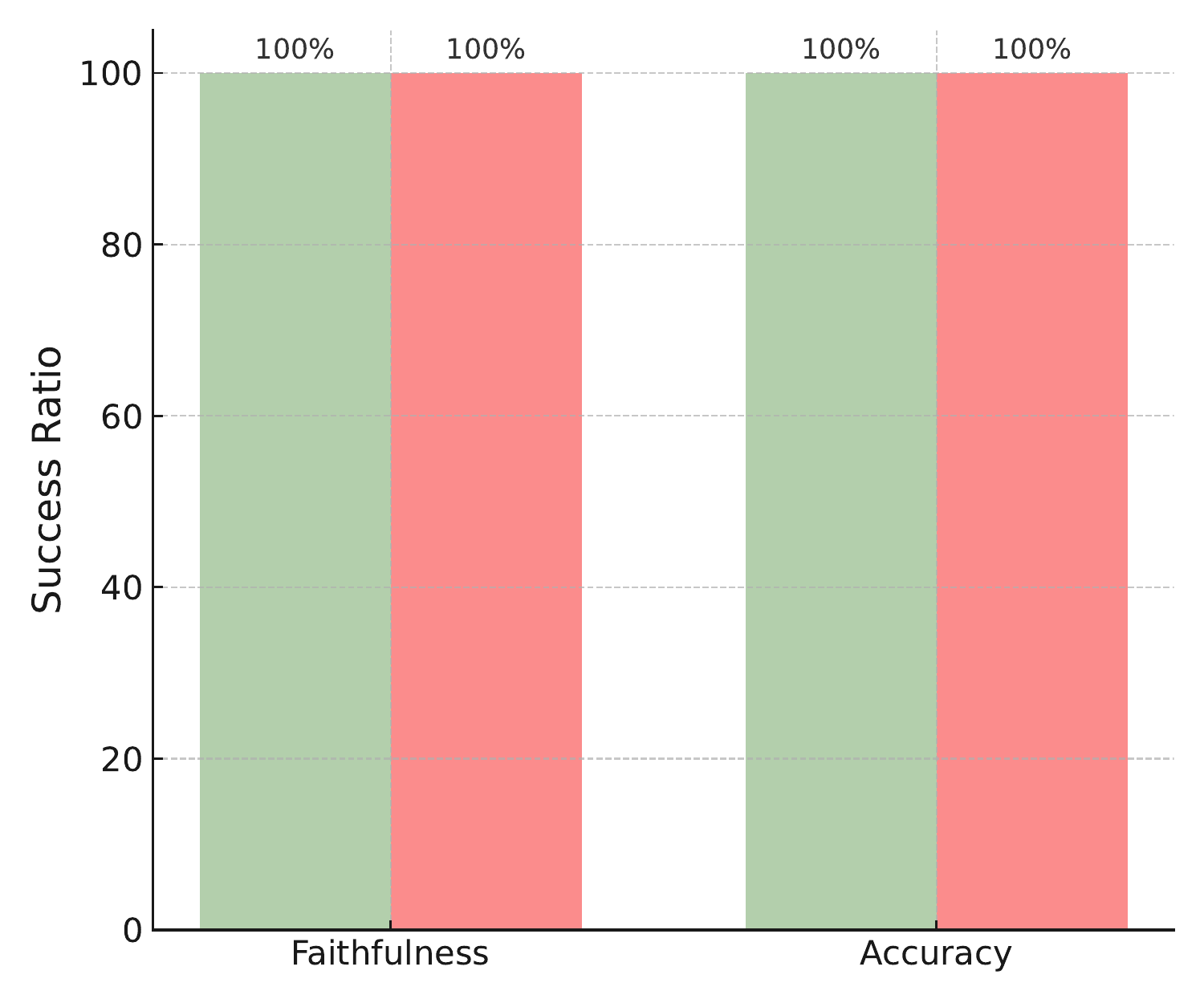}
        \caption{LM Debugger's accuracy in identifying the reliability of the model's predictions based on explanation quality.}
        \label{fig:lm_debugger}
\end{figure}

We demonstrate the effectiveness of the LM Debugger in helping users determine when to trust the model's predictions. We set a threshold score of 3 and focus on the accuracy-related dimensions, specifically ``Faithfulnes'' and ``Accuracy'', which serve as guides for assessing the LM's prediction accuracy.
Figure \ref{fig:lm_debugger} presents the percentage of cases where the LM Debugger correctly identifies the reliability of the model's predictions based on the explanation. The results show that our debugging process effectively assesses the dependability of the model's predictions based on the \methodname{} generated explanation, achieving 100\% accuracy in evaluating faithfulness and accuracy for both correct and incorrect predictions.

\section{Details of Element-graph} \label{sec:graph}
Due to space constraints in the main text, we provide a comprehensive description of the node and relations types, alongside the detailed equations for computing their embeddings. 

The node-type $u_i$ are the one-hot vectors of the node types. The type is according to the node's origin form, the input content $z$, question $\{q\}$, answer $\mathcal{A}$, or the node in the KG. 
The $u_i$ is transformed into an embedding through a linear transformation for subsequent calculations.

The relation type $r_{ij}$ is determined using pre-defined templates, which are employed to extract relations from the knowledge triples in the KG~\citep{feng2020scalable}. The embedding $\boldsymbol r_{ij}$ for the relation is computed for subsequent use: 
\begin{equation}
  \boldsymbol r_{ij} = f_{\zeta } (r_{ij}, u_{ij}) = f_{\zeta } (r_{ij}, u_{i}, u_{j}),
\end{equation}
where $f_{\zeta }$ is a two-layer MLP, $u_{ij}$ denotes the concatenation of $u_{i}$ and $u_{j}$.

The node score $v_{score}$ is subsequently utilized in its embedded form, calculated by:
\begin{equation}
    \boldsymbol v_{score} = f_{\rho }(v_{score})
\end{equation}
where $f_{\rho }$ is a two-layer MLP.

\section{Other Explanation Examples}\label{sec:appendix-example}
We demonstrate the complete explanation example of PathReasoner and ECQA in Table~\ref{app:example}. These methods exhibit in an unclear and intricate manner. Such explanations make it hard for humans to understand the decision-making process behind the model.

\begin{table}[htb]
  \centering
  \resizebox{0.9\columnwidth}{!}{%
    \begin{tabular}{@{}c|l@{}}
      \toprule
      \textbf{Input Questions}                          & \begin{tabular}[c]{@{}l@{}}Q: What is someone doing if he or she is sitting quietly and his or \\her eyes are moving?\\ A. reading B. meditate C. fall asleep D. bunk E. think\end{tabular}                                                                                                                                                                                                                                                                                                                 \\ \midrule
      \textbf{Label}                                    & A. reading                                                                                                                                                                                                                                                                                                                                                                                                                                                                                                                                                                                                                                                                                    \\
      \midrule
                                                        & \textbf{Explanation of Others}                                                                                                                                                                                                                                                                                                                                                                                                                                                                                                                                                                                                                                                                \\ \midrule
      \textbf{\makecell{Path-                                                                                                                                                                                                                                                                                                                                                                                                                                                                                                                                                                                                                                                                                                                           \\Reasoner }} & \begin{tabular}[c]{@{}l@{}}quietly [related to] quiet [at location] a library [used for] reading \\ eyes [used for] reading \\ eyes [form of] eye [related to] glasses [used for] reading \\ sitting [related to] sit [related to] relaxing [has subevent] reading \\\end{tabular} \\ \midrule
      \textbf{\makecell{ECQA \\}} & \begin{tabular}[c]{@{}l@{}}\textbf{Positive examples:} \\ - When we read, our eyes move. \\ - While reading, a person sits quietly, \\ \textbf{Negative examples: }\\ - While meditating, eyes don't move, eyes are closed, \\ - While sleeping, eyes are closed and they don't move, \\ - When a person bunks, he/she doesn't sit quietly, \\ - Eyes don't move when you think about something. \\ \textbf{Explanation:} \\ When we read, our eyes move. \\ While reading, a person sits quietly. \\ While meditating and sleeping, eyes don't move, eyes are closed. \\ When a person bunks, he/she doesn't sit quietly. \\ Eyes don't move when you think about something. \\\end{tabular} \\ \bottomrule
    \end{tabular}%
  }
  \caption{The complete explanation examples of PathReasoner and ECQA.}
  \label{app:example}
\end{table}

\section{Case Study} \label{app:case}

In this section, we will present more explanation examples generated by~\methodname{}, to demonstrate the effectiveness of our method. These questions are randomly selected from the dataset. By sharing these examples, we aim to highlight the broad applicability and robust performance of~\methodname{} across diverse contexts. These examples not only illustrate the adaptability of our approach but also substantiate its validity by demonstrating consistent results. The comparison examples are shown in Figure \ref{fig:correct_ex} and Figure \ref{fig:wrong_exam}.

\definecolor{green}{HTML}{B7C47D}
\definecolor{red}{HTML}{D46155}

\begin{figure}[htb]
\centering
\begin{tcolorbox}[colback=softGray, colframe=green, title=\ding{52} Explanation Example of Correct Prediction, 
coltitle=black,
fonttitle=\fontsize{8pt}{1pt}\selectfont, fontupper=\fontsize{5pt}{1pt}\selectfont,]
\textbf{Input:} 
The question is ``What is someone doing if he or she is sitting quietly and his or her eyes are moving?''\\
The Answer Options are:
A. Reading, B. Meditate, C. Fall Asleep, D. Bunk, E. Think

\textbf{Output (Model Prediction):} “A. Reading”

\textbf{Top-ranked reason-elements:}
1. Quiet chattering mind, 2. Not making sound, 3. Mind focuses, 4. Glasses for people with poor eyesight, 5. War
\vspace{2mm} 
\hrule 
\vspace{2mm} 
\textbf{Stage 1 - Explanation (why-choose):}
The model predicts "Reading" (A) as the activity someone is engaged in when sitting quietly with moving eyes, based on the key reason-elements. Here's how the reason-elements support this conclusion:\\
- \highlight{\textbf{Quiet Chattering Mind \& Not Making Sound:}} Indicate silent, internal focus, typical of reading, fitting the description of someone sitting quietly.\\
- \highlight{\textbf{Mind Focuses:}} Suggests active engagement with content, aligning with reading.\\
- \highlight{\textbf{Glasses for Poor Eyesight:}} Implies a visual task, supporting reading despite being an indirect clue.\\
% - \highlight{\textbf{War:}} Irrelevant to the decision-making process.

Excluding outliers, these elements collectively signal silent, visually focused activity, characteristic of reading.

\textbf{Stage 2 - Explanation (why-not-choose):}
The model's reasoning for making other options less likely is as follows:\\
- \highlight{\textbf{Meditate (B):}} Involves closed or unfocused eyes, unlike reading's active eye movement.\\
- \highlight{\textbf{Fall Asleep (C):}} Entails cessation of eye movement, not alignment with reading.\\
- \highlight{\textbf{Bunk (D):}} Lacks specific association with quiet, focused eye movement.\\
- \highlight{\textbf{Think (E):}} Doesn't necessitate visual engagement, different from reading.

Thus, only reading aligns with the criteria of silent, focused activity involving active visual engagement.

\vspace{2mm} 
\hrule 
\vspace{2mm} 
\textbf{LM Debugger:}

- \highlight{\textbf{Faithfulness:}} 4/5 - Well-aligned with the LM's data-driven reasoning.

- \highlight{\textbf{Completeness:}} 4/5 - Captures key reasons for "Reading" but may miss minor factors distinguishing it from similar activities.

- \highlight{\textbf{Minimality:}} 4/5 - Includes mostly relevant reasoning but is diluted by the unrelated "War."

- \highlight{\textbf{Accuracy:}} 4/5 - Accurately identifies "Reading" as the most likely activity, with minor issues in relevancy and nuanced differentiation.

- \highlight{\textbf{Advice for Improvement:}} Enhance the LM by broadening training data to cover more \textbf{physical activities} for accuracy, and fine-tuning with \textbf{detailed activity datasets} for sharper recognition.
\end{tcolorbox}
\caption{Explanation Example of Correct Prediction.}
\label{fig:correct_ex}
\end{figure}

\begin{figure}[htb]
\centering
\begin{tcolorbox}[colback=softGray, colframe=red, title=\ding{56} Explanation Example of Incorrect Prediction, fonttitle=\fontsize{8pt}{1pt}\selectfont, fontupper=\fontsize{5pt}{1pt}\selectfont,]

\textbf{Input:} 
The question is ``What do parents tell a child to do on the weekend?''\\
The Answer Options are:
A. Study, B. Begin school, C. Go out to play, D. Row boat, E. Clean room

\textbf{Output (Model Prediction):} “E. Clean room”

\textbf{Top-ranked reason-elements:}
1. Place Baseball, 2. Asleep, 3. Subway, 4. Ups, 5. Mentors
\vspace{2mm} 
\hrule 
\vspace{2mm} 
\textbf{Stage 1 - Explanation (why-choose):}
The model's prediction of "E. Clean room" as the task parents tell a child to do on the weekend is influenced by the ranked reasoning elements, each contributing to the rationale in abstract but supportive ways:

- \highlight{\textbf{Place Baseball:}} Suggests activities requiring cleanup, reinforcing the idea of tidying up spaces like a room.

- \highlight{\textbf{Asleep:}} Implies end-of-day tasks, where cleaning up could be a preparatory action before rest.

- \highlight{\textbf{Subway:}} Though less direct, could symbolize daily routines, pointing to regular tasks like room cleaning.

- \highlight{\textbf{Ups:}} Might relate to picking up or organizing items, aligning with cleaning activities.

- \highlight{\textbf{Mentors:}} Reflects the guiding role of parents in instilling responsibility, including maintaining cleanliness.

Together, these elements hint at themes of responsibility, routine, and organization, leading to the conclusion that cleaning a room is a fitting weekend task for a child from a parent's perspective.

\textbf{Stage 2 - Explanation (why-not-choose):}
The model's reasoning for making other options less likely is as follows:\\
- \highlight{\textbf{A. Study}} and \highlight{\textbf{B. Begin school:}} They're associated with cognitive tasks, diverging from the model's focus on physical organization and day-end routines.

- \highlight{\textbf{C. Go out to play:}} Suggests initiating a new activity, not related to the model's emphasis on tidiness and concluding the day with responsible actions.

- \highlight{\textbf{D. Row boat:}} Similar to "C," it indicates starting an activity rather than organizing or cleaning, moving away from the themes of responsibility towards personal spaces.

Thus, the model's logic, centered around ending day activities and maintaining personal spaces, naturally leads to "E. Clean room" as the most logical weekend task for a child, sidelining the other options due to their lack of connection to tidiness and routine responsibility.

\vspace{2mm} 
\hrule 
\vspace{2mm} 
\textbf{LM Debugger:}

- \highlight{\textbf{Faithfulness:}} 1/5 - Misrepresents LM reasoning, attributing abstract thought where only statistical patterns apply.

- \highlight{\textbf{Completeness:}} 2/5 - Omits explanation of LMs' data-driven nature, focusing instead on unrelated elements.

- \highlight{\textbf{Minimality:}} 1/5 -  Includes extraneous elements, veering away from the statistical basis of LM outputs.

- \highlight{\textbf{Accuracy:}} 1/5 - Anthropomorphizes LM's statistical analysis as abstract reasoning, misinterpreting its function.

- \highlight{\textbf{Advice for Improvement:}} Focus on diversifying training datasets with a wider range of language use related to \textbf{children's tasks}, improve decision-making accuracy with \textbf{targeted training on family and child-rearing contexts}.
\end{tcolorbox}
\caption{Explanation Example of Incorrect Prediction.}
\label{fig:wrong_exam}
\end{figure}

\section{Ablation Studies}\label{sec:ablation}
In this section, we examine the impact of different components of~\methodname{} on its performance. We evaluated the effects of the size of the LMs, knowledge components, and interpreting components using the CommonsenseQA IHdev and IHtest datasets. Tables~\ref{ab:PLM},~\ref{ab:model-only} and~\ref{ab:interpreting} summarize the ablation study results.

\begin{table}[h]
  \centering
  \resizebox{0.49\columnwidth}{!}{%
  \begin{tabular}{@{}rcc@{}}
    \toprule
    Method                       & IHdev-Acc. & IHtest-Acc. \\ \midrule
    RoBERTa w/o itp              & 68.63\%    & 64.54\%     \\
    RoBERTa-large w/o itp        & 73.05\%    & 71.96\%     \\
    RoBERTa-large + itp & \textbf{77.97\%}    & \textbf{77.31\%}     \\ \bottomrule
  \end{tabular}
  }
  \caption{Ablation study on the effect of interpreting component on model accuracy.}
  \label{ab:interpreting}
\end{table}

Table~\ref{ab:PLM} shows the impact of the size of LM on~\methodname{}. We evaluate the performance of LMs with two different sizes: 1) RoBERTa-large (with 340 million parameters) and 2) RoBERTa (with 110 million parameters). The results show that using a larger LM leads to significant improvement in performance, with an increase of 11.71\% and 14.30\% in model accuracy on the IHdev dataset and the IHtest dataset, respectively.

Table~\ref{ab:model-only} shows the impact of the knowledge component of~\methodname{}.
We compare the performance of the LM-only model with and without external knowledge from ConceptNet.
\textit{only} means we only use the LM to predict the answer. \textit{+ external knowledge} means the external knowledge is leveraged. We observe that incorporating external knowledge significantly improves the accuracy of the LM prediction, especially on the test set. With external knowledge, the model accuracy on IHdev and IHtest is increased by at least 3.69\% and 7.12\%, respectively.
% This shows that external knowledge plays an important role in enhancing the reasoning ability of the model.

\begin{table}[h]
\centering
\resizebox{0.49\columnwidth}{!}{
\begin{tabular}{@{}rcc@{}}
    \toprule
    LM                    & IHdev-Acc. & IHtest-Acc. \\ \midrule
    RoBERTa               & 66.26\%    & 63.01\%     \\
    RoBERTa-large (final) & \textbf{77.97\%}    & \textbf{77.31\%}     \\\bottomrule
  \end{tabular}
  }
  \caption{Ablation study on the effect of LM size on model accuracy.}
  \label{ab:PLM}
\end{table}

In Table~\ref{ab:interpreting}, we analyze the impact of the interpreting component on LM performance. \textit{w/o itp} indicates that the interpreting component was not incorporated in the prediction, whereas the \textit{+ itp} indicates its presence. We observe that removing the interpreting component leads to a clear decrease in accuracy by at least 4.92\% and 5.35\% on IHdev and IHtest, respectively. Furthermore, comparing the results of \textit{RoBERTa-large only}, \textit{RoBERTa-large + itp}, and \textit{final}, we find that the interpreting component has a greater impact on accuracy than the other components.

\begin{table}[htb]
\centering
\resizebox{0.49\columnwidth}{!}{
\begin{tabular}{@{}rcc@{}}
    \toprule
    Method                        & IHdev-Acc. & IHtest-Acc. \\ \midrule
    RoBERTa only                  & 62.65\%    & 60.27\%     \\
    RoBERTa-large only            & 74.28\%    & 70.19\%     \\
    RoBERTa-large + external knowledge & \textbf{77.97\%}    & \textbf{77.31\%}     \\ \bottomrule
  \end{tabular}
  }
  \caption{Ablation study on the effect of knowledge component on model accuracy.}
  \label{ab:model-only}
\end{table}

The ablation highlights the positive contributions of each component of \methodname{}. Specifically, we find that the interpreting component plays a crucial role in enhancing model accuracy and generalizability on unseen questions.

\subsection{Results of Different Generators} \label{appendix:generators}

\begin{figure*}[htb]
  \begin{center}
    %\framebox[4.0in]{$\;$}
    %\fbox{\rule[-.5cm]{0cm}{4cm} \rule[-.5cm]{4cm}{0cm}}
    \includegraphics[width=1\textwidth]{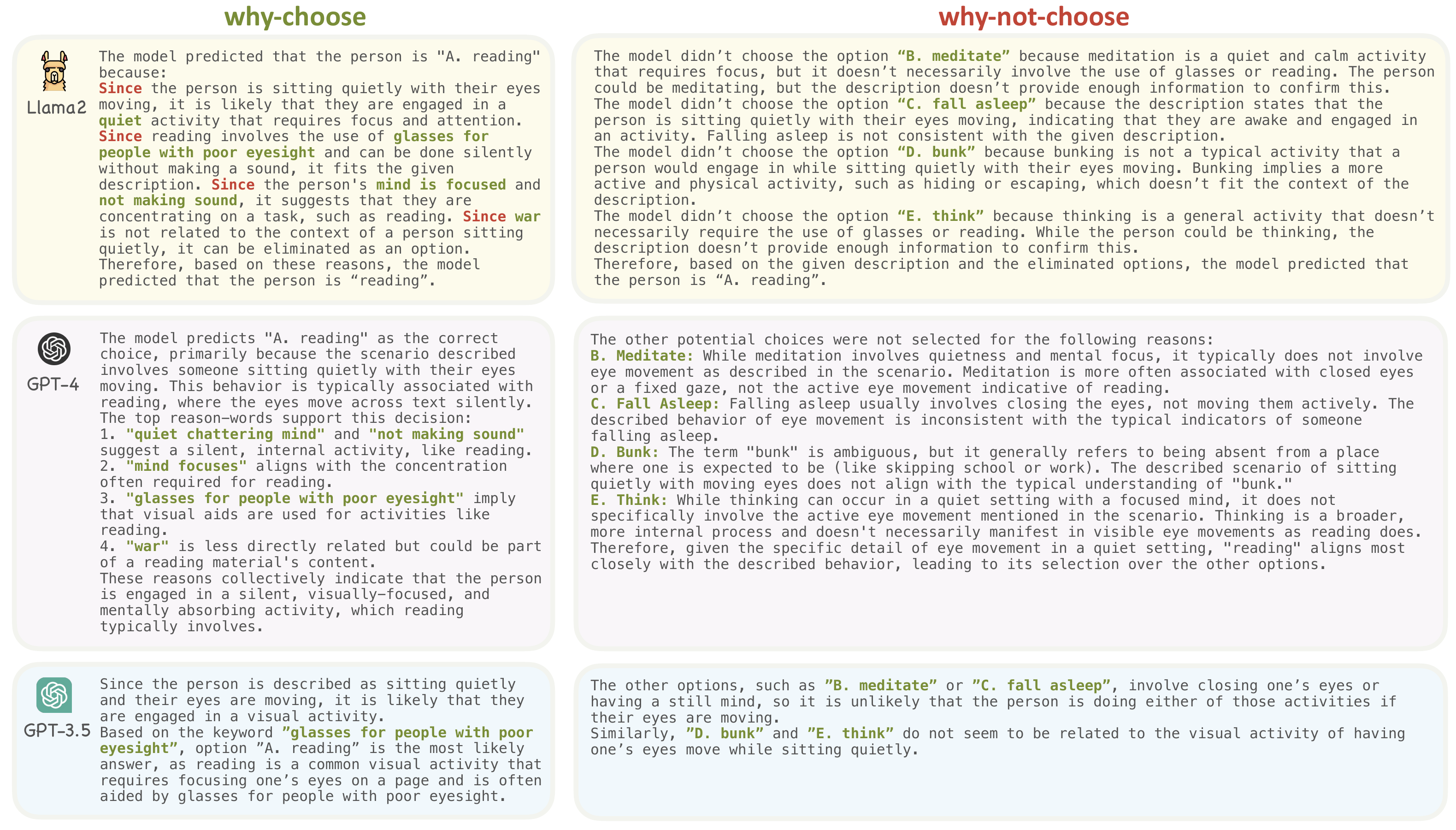}
  \end{center}
  \caption{The \textit{why-choose} and \textit{why-not-choose} explanations generated by Llama-2-70B, GPT-4 and GPT-3.5. The semantic meanings remain consistently aligned among the explanations generated by the three models.}
  \label{fig:appendix-exam}
\end{figure*}

In this section, we present a comprehensive analysis of the results from different explanation generators: Llama-2-70B, GPT-4-turbo, and GPT-3.5-turbo. We focus on evaluating how each generator interprets and translates the model's decision-making process into human-understandable explanations.

The complete experimental results are presented in Figure \ref{fig:appendix-exam}, where all experiments are conducted under the same settings. The question is collected randomly: 
\begin{itemize}
    \item Question: What is someone doing if he or she is sitting quietly and his or her eyes are moving?
    \item Answer Choices: A. reading, B. meditate, C. fall asleep, D. bunk, E. think.
    \item Model Prediction: A. reading
    \item Ground-truth Answer: A. reading
\end{itemize}

We utilize RoBERTa-large as the LM $f_{LM}$ for this experiment. The $f_{LM}$ correctly predicts the answer as ``A. reading''. Our extracted \textit{reason-elements} are: 1. quiet chattering mind, 2. not making sound, 3. mind focuses, 4. glasses for people with poor eyesight, 5. war.

To further quantify the semantic similarity between explanations of Llama-2, GPT-4, and GPT-3.5, we employ GPT-4 to generate similarity scores. GPT-4's advanced language comprehension abilities make it well-suited for this task, offering a human-like understanding of textual content. The scores reflect the degree of alignment in content among the explanations. The score is on a scale from 0 to 1, where 1 is very similar and 0 is not similar at all.

\begin{figure}[h]
  \begin{center}
    \includegraphics[width=0.45\textwidth]{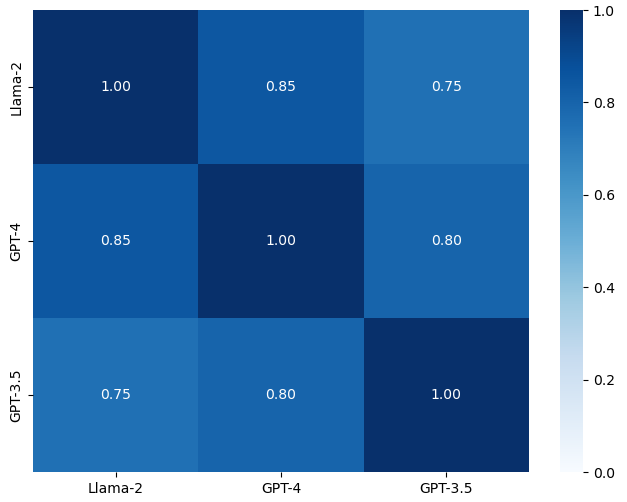}
  \end{center}
  \caption{Heatmap of Similarity Scores for Llama-2, GPT-4, and GPT-3.5: Their generated explanations show high consistency in terms of semantic meaning. }
  \label{fig:appendix-heat}
  \vspace{-10pt}
\end{figure}

\textbf{Llama-2 vs. GPT-4:}

Similarity: Both explanations align in focusing on the 'reading' activity, referencing quiet sitting, eye movement, and glasses use.

Similarity Score: 0.85/1 - High similarity in core reasoning and conclusion.

\textbf{Llama-2 vs. GPT-3.5:}

Similarity: Both identify the person as engaged in reading, noting quiet sitting and glasses use.

Similarity Score: 0.75/1 - Similar in conclusion and main points, but GPT-3.5 provides more concise content.

\textbf{GPT-4 vs. GPT-3.5:}

Similarity: Agreement in the conclusion of ``reading'', common elements include quiet posture, eye movement, and glasses use.

Similarity Score: 0.80/1 - Similar key conclusions and elements, but GPT-4 includes more detail.

We illustrate the similarity scores in Figure \ref{fig:appendix-heat}. The color intensities represent the degree of similarity, with darker tones indicating higher congruence. It shows their generated explanations align in semantic meaning. 

% Despite variations in style and detail, the fundamental meanings are consistent across all generators. This consistency highlights the effectiveness of our approach in preserving the accuracy of the explanations.

The explanations generated by the three models are largely consistent in semantic meaning, demonstrating that under our constrained prompt instruction, these models primarily functioned as ``translators''. They convert the reasoning process into human-understandable language. However, it is important to note that the capability of the generator influenced the readability of the explanations. For instance, Llama-2 tends to produce more repetitive language (in red), while GPT-3.5-turbo and GPT-4 show consistency and conciseness. Based on these observations, we recommend using GPT-3.5-turbo or GPT-4 as the explanation generator for optimal clarity.

\section{Material for User Perception Perspective Evaluation}
\subsection{Details of Evaluation Metrics}
\paragraph{Overall Quality} This criterion assesses the effectiveness of the explanations in making the LMs' decision processes understandable to users, providing a comprehensive measure of the explainability's efficacy.

\paragraph{Understandability} This metric evaluates the clarity and coherence of the explanations, determining how easily users can comprehend the model's outputs and underlying reasoning.

\paragraph{Trustworthiness} This assesses users' confidence in the model's outputs and explanations, examining whether the explanations are perceived as reliable, credible, and based on sound logic.

\paragraph{Satisfaction} This captures users' overall contentment with the explanations, considering whether the outputs meet their expectations in terms of quality, relevance, and usefulness.

\paragraph{Detail Sufficiency} This examines whether the explanations provide an adequate level of detail, ensuring they are sufficiently descriptive and comprehensive to fully address the question or task at hand.

\paragraph{Completeness} This evaluates the extent to which the explanations cover the model's decision-making process, verifying that all essential aspects of the model's reasoning are addressed.

\paragraph{Accuracy} In addition to objective measures, this involves a subjective assessment of whether the explanations align with users' knowledge or expectations, confirming the factual correctness and contextual appropriateness of the model's outputs and explanations.

\subsection{Details of Automated Evaluation}\label{appendix:auto-eval}

In this section, we describe the process of using automated evaluators to evaluate the quality of our explanations. We start by randomly selecting an example to demonstrate the evaluation procedure. The evaluators provide their scores and reasons for their ratings. This allows us to verify the validity of their assessments and better understand the basis of their evaluations. We use GPT-3.5-turbo and GPT-4 as our automated evaluators.

\textbf{Question:} What might happen when not being careful setting a cup on a table?

A. picks up

B. noise

C. decorating the table

D. spillage

E. clutter

\textbf{LM predicted answer:}  D. spillage

\textbf{Ranked Reason-elements:} 1. cold day, 2. careless, 3. past can, 4. quite possible, 5. written document.

\textbf{\textit{Why-choose} explanation:} Since the person is being careless and it's quite possible that they've spilled things in the past, the model predicts that there is a high likelihood of spillage when not being careful setting a cup on a table. 

\textbf{\textit{Why-not-choose} explanation:} The other potential choices, such as picks up, decorating the table, and clutter, were not predicted because those words were not mentioned as significant reasons by the model. Additionally, there was no indication in the reason-words to suggest that those actions would occur as a result of not being careful when setting a cup on a table. The word noise was also not predicted because it does not directly relate to the action of setting a cup on a table, nor were any reason-words related to it.

\textbf{Prompt: }Suppose you are an AI expert. We now require your assistance in evaluating the content validity of explanations that why a language model made certain decisions. Please adhere to our grading rubric, using ``1'' to represent ``Disagree'', ``2'' for ``Neutral'', and ``3'' for ``Agree''. Your grading should be based on the following 8 questions: 

1. This is a good explanation; 

2. I understand this explanation of how the AI model works; 

3. I trust this explanation of how the AI model works; 

4. This explanation of how the AI model works is satisfying; 

5. This explanation of how the AI model works provides sufficient detail; 

% 6. This explanation of how the AI model works contains irrelevant details; 

6. This explanation of how the AI model works seems complete; 

7. This explanation of how the AI model works is accurate.

\textbf{Evaluation results: }

\textbf{GPT-3.5-turbo:} 

1. This is a good explanation: 3 (Agree)

2. I understand this explanation of how the AI model works: 3 (Agree)

3. I trust this explanation of how the AI model works: 3 (Agree)

4.  This explanation of how the AI model works is satisfying: 3 (Agree)

5. This explanation of how the AI model works has sufficient detail: 3 (Agree)

% 6. This explanation of how the AI model works contains irrelevant details: 2 (Neutral)

6. This explanation of how the AI model works seems complete: 3 (Agree)

7. This explanation of how the AI model works is accurate: 3 (Agree) 

Reason: The explanation provided is clear and understandable, providing a valid reasoning for why the predicted answer is ``spillage''. The mention of being careless and the likelihood of spillage based on past experiences supports the prediction that not being careful when setting a cup on a table could result in spillage. However, some of the reason-elements, such as cold day, quite possible, and written document, do not seem directly relevant to the context and do not contribute significantly to the explanation, resulting in a neutral rating for question 6. Nonetheless, the overall explanation is satisfactory and accurate given the available information.

\textbf{GPT-4:} 

1. This is a good explanation: 3 (Agree)

2. I understand this explanation of how the AI model works: 3 (Agree)

3. I trust this explanation of how the AI model works: 3 (Agree)

4. This explanation of how the AI model works is satisfying: 3 (Agree)

5. This explanation of how the AI model works has sufficient detail: 3 (Agree)

% 6. This explanation of how the AI model works contains irrelevant details: 2 (Neutral)

6. This explanation of how the AI model works seems complete: 3 (Agree)

7. This explanation of how the AI model works is accurate: 3 (Agree)

Reason: While the reason-elements (``cold day'', ``careless'', ``past can'', ``quite possible'', ``written document'') don't seem to have a clear relation to the specific context of the question, the explanation itself is still understandable, providing clear reasoning for why the model chose ``spillage'' as the most likely outcome and why the other options were less likely. The explanation is detailed, complete, and aligns with common sense and real-world expectations about what might happen when someone is not careful while setting a cup on a table.

\end{document}